\documentclass[11pt,a4paper]{article}
\usepackage[hyphens]{url}
\usepackage[hyperref]{emnlp2020}
\usepackage{times}
\usepackage{latexsym}
\usepackage{colortbl}
\usepackage{array,multirow,graphicx}

\usepackage{microtype}

\aclfinalcopy 

\usepackage[textsize=tiny,disable]{todonotes}
\usepackage{booktabs}
\usepackage{amsmath,amssymb}
\usepackage{enumitem}

\newcommand{\en}{\texttt{en}}
\newcommand{\tr}{\texttt{tr}}
\newcommand{\ru}{\texttt{ru}}
\newcommand{\ka}{\texttt{ka}}
\newcommand{\pl}{\texttt{pl}}
\newcommand{\de}{\texttt{de}}

\newcommand{\se}[1]{\textcolor{black}{#1}}

\title{How to Probe Sentence Embeddings in Low-Resource Languages: \\ On Structural Design Choices for Probing Task Evaluation}

\author{Steffen Eger, Johannes Daxenberger, Iryna Gurevych\\
    Computer Science Department, Technische Universit\"at Darmstadt, Germany \\
     \texttt{{eger}@aiphes.tu-darmstadt.de},\\ \texttt{\{daxenberger,gurevych\}@ukp.informatik.tu-darmstadt.de} \\
 }

\date{}

\begin{document}
\maketitle
\begin{abstract}
Sentence encoders map sentences 
to real valued vectors 
for use in downstream applications. 
To peek into these representations---e.g., to increase interpretability of their results---probing tasks 
have been designed which query them for linguistic knowledge. However, designing probing tasks for lesser-resourced languages is tricky, because these often lack large-scale annotated data or (high-quality) dependency parsers as a prerequisite of probing task design in English. 
To investigate how to probe sentence embeddings in such cases, 
we investigate sensitivity of probing task results to structural design choices, conducting the first such large scale study. 
We show that design choices like  
size of the annotated probing dataset and type of classifier used for evaluation  
do (sometimes substantially) influence probing outcomes. 
We then probe embeddings in a multilingual setup with design choices that lie 
in 
a `stable region',  
as we identify for English, and find that results on English do not transfer to other languages. 
Fairer and more comprehensive sentence-level 
probing evaluation 
should thus be carried out on multiple languages in the future. 
\end{abstract}

\section{Introduction}

\se{
Sentence embeddings (a.k.a.\ sentence encoders)
have become ubiquitous in NLP \citep{Kiros.2015,Conneau.2017}, extending 
the concept of word embeddings to the sentence level.}  
In the context of recent efforts to open the black box of deep learning models and representations \citep{ws-2019-2019-acl}, it has also become fashionable to \emph{probe} sentence embeddings for the linguistic information signals they contain \citep{Perone.2018}, as this may not 
be clear from their performances in downstream tasks. Such probes are linguistic micro tasks---like detecting the length of a sentence or its dependency tree depth---that have to be solved by a 
classifier using given representations. 

The majority of approaches for probing sentence embeddings 
target English, 
but recently some works have also addressed other languages such as Polish, Russian, or Spanish in a multi- and cross-lingual setup 
\cite{Krasnowska.2019,ravishankar-etal-2019-probing}. Motivations for considering a multi-lingual analysis include knowing whether findings from English transfer to other languages and determining a universal set of probing tasks that suits multiple languages, e.g., with richer morphology and freer word order.  

Our work is also inspired by probing sentence encoders in multiple (particularly low-resource) languages.  
We are especially interested in the 
formal structure of probing task design in this context.
Namely, when designing probing tasks for low-resource languages, some questions arise naturally that are less critical in English. One of them is the 
size
of training data for probing tasks, as this training data typically 
needs to be (automatically or manually) annotated, 
an inherent obstacle in low-resource settings.\footnote{The main issue is that high-quality dependency parsers, as required for standard probing tasks, exist only for a handful of languages. E.g., UDPipe \citep{straka-2018-udpipe} is available for only about 100 languages, and performance scores for some of these are considerably below those of English \citep{straka-2018-udpipe}.}

\begin{table}[]
    \centering
    {\small
    \begin{tabular}{cc|cccc}
         \toprule
         & \multicolumn{5}{c}{\textbf{classifier}} \\
        &      & LR & MLP & NB & RF \\ \midrule
        \parbox[t]{2mm}{\multirow{3}{*}{\rotatebox[origin=c]{90}{\textbf{size}}}} & High &  \cellcolor{blue!25}(A,B,C) & \cellcolor{blue!25}(A,B,C) & (C,A,B) & (C,B,A) \\
    
    & Mid &   (A,C,B) & (C,B,A) & \cellcolor{blue!25}(A,B,C) & (C,B,A)\\
        & Low &   \cellcolor{blue!25}(A,B,C) & (B,A,C) & (B,C,A) & \cellcolor{blue!25}(A,B,C) \\
         \bottomrule
    \end{tabular}
    \caption{Schematic illustration of our concept of stability across two dimensions (classifier and training size). Here, three encoders, dubbed A,B,C, are ranked. The region of stability is given by those settings that support the majority ranking of encoders, which is  A$\succ$B$\succ$C.}
    \label{table:stability}
    }
\end{table}

Thus, 
at first, 
we ask for the training data size required for obtaining reliable probing task results. 
This question  
is also relevant 
for English: 
on the one hand, \citet{Conneau.2018a} claim that training data for a probing task should be plentiful, as otherwise (highly parametrized) classifiers on top of representations may be unable to extract the relevant information signals; 
on the other hand, \citet{hewitt-liang-2019-designing} note 
that a sufficiently powerful classifier with enough training data can in principle 
learn any task, 
without this necessarily allowing to conclude that the representations 
adequately store  the linguistic signal under scrutiny.
Second, we 
ask 
how stable probing task results are across  different classifiers (e.g., MLP vs.\ Naive Bayes). This question is closely related to the question about size, since different classifiers have different sensitivities to  
data size; especially deep models are claimed to require more training data.

We evaluate 
the sensitivity of probing task results to 
the 
two outlined parameters---which are \emph{mere} machine learning design choices  
that do not affect 
the linguistic content 
stored in the sentence representations under scrutiny---and then determine a 
`region of stability' for English (\en), where outcomes are predicted to be similar for the majority of parameter choices. Table  \ref{table:stability} illustrates this. 
Using parameter choices within our region of stability, we turn to three lower-resource languages, 
\emph{viz.}: Turkish (\tr), Russian (\ru), and Georgian (\ka).
\tr{} is a Turkic language written in Latin script which 
makes exhaustive use of agglutination.
\ru{} is a Slavic language written in Cyrillic script 
characterized by strong inflection and rich morphology.
\ka{} is a South-Caucasian language using its own script called Mkhedruli.
It makes use of both agglutination as well as inflection. 
For these, our main research questions are whether probing task results transfer from English to the other languages. 

Overall, our research questions are: 
\begin{itemize}[topsep=3pt,itemsep=0pt,leftmargin=*]
\item \textbf{(i)} How reliable are probing task results across machine learning design choices?
    \item \textbf{(ii)} 
    Will encoder performances correlate across languages,  
    even though the languages and their linguistic properties may differ? 
    \item \textbf{(iii)} 
    Will probing task performances  
    correlate across languages?
    \item \textbf{(iv)}  
    Will the correlation between probing and downstream tasks 
    be the same across languages? 
\end{itemize}

These questions are important because they indicate whether or not probing tasks (and their relation to downstream tasks) have to be re-evaluated in languages other than \en{}.\footnote{Code and data are available from \url{https://github.com/UKPLab/conll2020-multilingual-sentence-probing}.} 

\section{Related work}

Our goal 
is to probe for sentence-level linguistic knowledge encoded in sentence embeddings \cite{Perone.2018} in a  
multilingual setup 
which marginalizes out the effects of probing task design choices when comparing sentence representations.

\textbf{Sentence embeddings} have become 
central 
for representing texts beyond the word level, e.g., in small data scenarios, where it is difficult to induce good higher-level text representations from word embeddings \citep{subramanian2018learning} or for clustering or text retrieval applications \citep{Reimers.2019}. 
To standardize the comparison of sentence embeddings, \newcite{conneau-kiela-2018-senteval} proposed the SentEval framework for evaluating the quality of sentence embeddings on a range of downstream and 10 probing tasks.

\textbf{Probing tasks} are used to introspect embeddings for linguistic knowledge, by taking ``probes'' as dedicated syntactic or semantic micro tasks
\cite{kohn-2016-evaluating}.
As opposed to an evaluation in downstream applications or benchmarks like GLUE \cite{wang-etal-2018-glue}, 
probing tasks  
target very specific linguistic knowledge which may otherwise be confounded in downstream applications. 
Since they are artificial tasks, they can also be better controlled for to avoid dataset biases and artifacts. 
Probing is typically either executed on type/token (word) \citep{tenney2018what} or sentence level \citep{Adi.2017}.   
For sentence level evaluation, 
SentEval thus far only includes \en{} data. 
Each probing task in SentEval is balanced and has 100k train, 10k dev, and 10k test instances.  
The effects of these design choices are  unclear, which is why our work addresses their influence systematically.

In the \textbf{multilingual} setting, 
\newcite{DBLP:journals/corr/abs-1903-09442} propose 15 token and type level probing tasks. 
Their probing task data is sourced from UniMorph 2.0 \cite{kirov-etal-2018-unimorph}, Universal Dependency treebanks \cite{mccarthy-etal-2018-marrying} and Wikipedia word frequency lists.
To deal with lower-resourced languages, they only use 10K samples per probing task/language pair (7K/2K/1K for train/dev/test) and exclude task/language pairs for which this amount cannot be generated.
Their final experiments are carried out 
on 
five languages (Finnish, German, Spanish, \ru{}, \tr{}), 
for which enough training data is available. 
They find that for morphologically rich (agglutinative) languages, several probing tasks positively correlate with downstream applications. Our work also investigates correlation between probing and downstream performance, but we do so on sentence level. 

On sentence level, \newcite{ravishankar-etal-2019-probing} 
train an InferSent-like encoder \citep{Conneau.2017} on \en{} and map this encoder to four languages (\ru{}, French, German, Spanish) using parallel data.
Subsequently, they 
probe the encoders on the probing tasks proposed by \newcite{Conneau.2018a} using Wikipedia data for each language, 
with the same size of probing task data as in SentEval, i.e., 100k/10k/10k for train/dev/test.  
Their interest is in whether probing tasks results are higher/lower compared to \en{} scores. They find  particularly the \ru{} probing scores to be low, which they speculate to be 
an artifact of cross-lingual word embedding induction and the language distance of \ru{} to \en{}.  
In contrast to us, their focus is particularly on the effect of transferring sentence representations from \en{} to other languages. 
The problem of such an analysis is that results may be
affected 
by the nature of the cross-lingual mapping techniques. 
\newcite{Krasnowska.2019} 
probe sentence encoders in \en{} and Polish (\pl{}). They use tasks defined in \newcite{Conneau.2018a} but slightly modify them (e.g., replacing dependency with constituency trees), reject some tasks (Bigram-Shift, as 
word order may play a minor role in \pl), and add two new tasks (Voice and Sentence Type). Since \pl{} data is less abundant, 
they shrink the size of the 
\pl{} datasets to 75k/7.5k/7.5k for train/dev/test and, for consistency, do the same for \en.
They extract probing datasets from an \en{}-\pl{} parallel corpus using COMBO for dependency parsing \citep{Rybak.2018}. 
They  
find that \en{} and \pl{} probing results mostly agree, i.e., encoders store the same linguistic information across the two languages. 
\section{Approach}

\begin{table*}[!htb]
    \centering
    {\footnotesize
    \begin{tabular}{l|llr} \toprule
         \textbf{Task} & \textbf{Description} & \textbf{Example} \\
         \midrule 
         Bigram Shift & Whether two words in a sentence are inverted & This is my Eve Christmas. $\longrightarrow$ True\\
         Tree Depth & Longest path from root to leaf in constituent tree & ``One hand here , one hand there , that 's it'' $\longrightarrow$ 5 \\
         Length & Number of tokens & I like cats $\longrightarrow$ 1-4 words\\ 
         Subject Number & Whether the subject is in singular or plural & They work together $\longrightarrow$ Plural\\
         Word Content & Which mid-frequency word a sentence contains & Everybody should step back $\longrightarrow$ everybody\\
         Top Constituents & Classific. task where classes are given by 19 most & Did he buy anything from Troy $\longrightarrow$ VDP\_NP\_VP \\
         & common top constituent sequences in corpus\\ \midrule
         Voice & Whether sent.\ contains a passive construct & He likes cats $\longrightarrow$ False\\
         SV-Agree & Whether subject and verb agree & They works together $\longrightarrow$ Disagree\\
         SV-Dist & Distance between subject and verb & The delivery was very late $\longrightarrow$ 1\\
         \bottomrule
    \end{tabular}
    }
    \caption{Probing tasks, their description and illustration. Upper tasks are defined as in SentEval.}
    \label{table:probing}
\end{table*}

In the absence of ground truth, 
our main interest is in 
a `stable' structural setup for probing task design---with the end goal of applying this design to multilingual probing analyses (keeping their restrictions, e.g., small data sizes, in mind). 
To this end, we consider a 2d  
space $\mathcal{X}$ comprising probing data size and classifier choice  
for probing tasks.\footnote{We also looked at further parameters, in particular, the class (im)balances of training datasets. Details and results can be found in the appendix. 
Since, however, their influence seemed to be less critical and 
an increased search space would blow up computational costs, we decided to limit our investigation to the described dimensions.} 
For a selected set of points $p_0,p_1,\ldots$ in $\mathcal{X}$, we evaluate all our encoders on $p_i$, and determine the `outcomes' $O_i$ (e.g., ranking) of the encoders at $p_i$. 
We consider a setup $p_i$ as stable if 
outcome $O_i$ is shared by a majority of other settings $p_j$. 
This can 
be considered a region of agreement, similarly to 
inter-annotator agreement \cite{artstein:2008}. 
In other words, we identify `ideal' test conditions by minimizing the influence of parameters $p_i$ on the outcome $O_i$. Below, we will approximate these intuitions using correlation. 

\subsection{Embeddings}
We consider two types of sentence encoders, \textbf{non-parametric methods} which combine word embeddings in elementary ways, without training; and \textbf{parametric methods}, which tune parameters on top of word embeddings. 
As non-parametric methods, we consider: (i) average word embeddings as a  
popular baseline,  
(ii) the concatenation of average, min and max pooling (\emph{pmeans}) \citep{Rueckle.2018}; and Random LSTMs \citep{Conneau.2017,Wieting.2019}, which feed word embeddings to 
randomly initialized LSTMs, then apply a pooling operation across time-steps.  
As parametric methods, we consider: InferSent \citep{Conneau.2017}, which induces a sentence representation by learning a semantic entailment relationship between two sentences; QuickThought \citep{Logeswaran.2018} 
which reframes the popular SkipThought model \citep{Kiros.2015} in a classification context; 
LASER \citep{ArtetxeS19} derived from massively multilingual machine translation models, and BERT base \citep{devlin-etal-2019-bert}, where we average 
token embeddings of the last layer for a sentence representation.
Dimensionalities of encoders are listed in the appendix.

\subsection{Probing Tasks}
Following \citet{Conneau.2018a}, we consider the following probing tasks: \textbf{BigramShift} (\en, \tr, \ru, \ka), \textbf{TreeDepth} (\en), \textbf{Length} (\en, \tr, \ru, \ka), \textbf{Subject Number} (\en, \tr, \ru), 
\textbf{WordContent} (\en, \tr, \ru, \ka), and 
\textbf{TopConstituents} (\en). 

We choose Length, BigramShift and WordContent because they are unsupervised tasks that require no labeled data and can thus be easily implemented across different languages---they also represent three different types of elementary probing tasks: surface, syntactic and semantic/lexical. We further include Subject Number across all our languages 
because 
number marking is extremely common across languages and it is comparatively easy to identify. 
We adopt \textbf{Voice} (\en, \tr, \ru, \ka)  from \citet{Krasnowska.2019}. 
For \en, we additionally evaluate on TreeDepth and TopConstituents as hard syntactic tasks. 
We add two tasks not present in the canon of probing tasks given in  
SentEval: 
\textbf{Subject-Verb-Agreement (SV-Agree)} (\en, \tr, \ru, \ka) and \textbf{Subject-Verb-Distance (SV-Dist)} (\en, \tr, \ru). We probe representations for these properties because we suspect that agreement between subject and verb is a difficult task which requires inferring a relationship between pairs of words which may stand in a long-distance relationship \citep{Gulordava.2018}. Moverover, we assume this task to be particularly hard in morphologically rich and word-order free languages, thus it could be a good predictor for performance in downstream tasks. 

To implement the probing tasks, for \en{}, we use the probing tasks datasets defined in \citet{conneau-kiela-2018-senteval} and we apply spaCy\footnote{\url{https://spacy.io}} to sentences extracted from Wikipedia for the newly added probing tasks Voice  
and SV-Agree. 
For \tr{}, \ru{}, and \ka{}, we do not rely on dependency parsers 
because of quality issues and unavailability for \ka{}. 
Instead, for \tr and \ru, we use information from Universal Dependencies (UD) \citep{nivre-etal-2016-universal}.   
 E.g., for SV-Dist, we 
determine the dependency distance between the main verb and the subject from UD. Instead of 
the exact distances, we predict binned classes: [1], [2,4], [5,7], [8,12], [13,$\infty$).  
For \ka, we use data and grammatical information from the Georgian National Corpus (GNC)\footnote{\url{https://clarino.uib.no/gnc}}.
We could not implement SV-Dist for \ka, due to missing dependency information in the GNC. 
For the same reason, we omit Subject Number for \ka{}.

For 
SV-Agree, we create a list of 
frequently occurring verbs together with their corresponding \emph{present tense conjugations} for all 
involved languages including English. 
We check each individual candidate  
sentence from Wikipedia 
for the presence of a verb form 
in the list. If no word is present, we exclude the sentence from
consideration. 
Otherwise, we randomly replace the verb form by a different conjugation in 50\% of the
cases.

An overview of the probing tasks, along with descriptions and examples, is given in Table \ref{table:probing}.

\subsection{Downstream Tasks}

In addition to  
probing tasks, we test the embeddings in downstream applications.  
Our focus is on a diverse set of high-level sentence classification tasks.
We choose \textbf{Argument Mining}, \textbf{Sentiment Analysis} and \textbf{TREC question answering}. 
Required training data for languages other than \en{} has been machine translated using Google Translate\footnote{\url{http://translate.google.com}} for Argument Mining and TREC.
\footnote{To estimate the quality of the machine translation,  
we measured 
its performance on parallel data. 
Details 
can be found in the appendix. \se{While the machine translation is generally of acceptable quality, we cannot exclude the possibility that it may effect some of our downstream tasks results reported below.}}
Statistics for all datasets are reported in Table \ref{table:multilingual_probing}. 

\paragraph{Argument Mining (AM)}
AM is 
an emergent NLP task requiring sophisticated reasoning capabilities.
We reuse the sentence-level argument (stance) detection dataset by \citet{Stab.2018}, which labels sentences extracted from web pages as pro-, con-, or non-arguments for eight different topics.
A sentence only qualifies as pro or con argument when it both expresses a stance towards the topic and gives a reason for that stance.
The classifier input is a concatenation of the sentence embedding and the topic encoding. 
In total, there are about 25,000 sentences.

\paragraph{Sentiment Analysis}
As opposed to AM, sentiment analysis only determines the opinion flavor of a statement.  
Since sentiment analysis is a very established NLP task, we did not machine translate \en{} training data, but used original data for \en{}, \ru{} and \tr{} and created a novel dataset for \ka{}.
For \en{}, we use the US Airline Twitter Sentiment dataset, consisting of 14,148 tweets labeled in three sentiment classes\footnote{\url{https://www.kaggle.com/crowdflower/twitter-airline-sentiment}}.
For \tr{}, we took the Turkish Twitter Sentiment Dataset with 6,172 examples and three classes\footnote{\url{https://github.com/hilalbenzer/turkish-sentiment-analysis}}.
For \ru{}, we used the Russian Twitter Corpus (RuTweetCorp), which we reduced to 30,000 examples in two classes.\footnote{\url{http://study.mokoron.com/}}
For \ka{}, we followed the approach by \newcite{choudhary-etal-2018-twitter} and crawled sentiment flavored tweets in a distant supervision manner.
Emojis were used as distant signals to indicate sentiment on preselected tweets from the Twitter API.
After post-processing, we were able to collect 11,513 Georgian tweets in three sentiment classes.
The dataset will made available publicly, including more details on the creation process.

\paragraph{TREC Question Type Detection}

Question type detection is an important part of Question-Answering systems. 
The Text Retrieval Conference (TREC) dataset consists of a set of questions 
labeled with their respective question types (six labels including e.g. ``description'' or ``location'') and is part of the SentEval benchmark \cite{conneau-kiela-2018-senteval}.
We used the data as provided in SentEval,  
yielding 5,952 instances.

\section{Experiments}

\paragraph{Experimental Setup}
To the SentEval toolkit \citep{conneau-kiela-2018-senteval}, which addresses both probing and downstream tasks and offers Logistic Regression (LR) and MLP classifiers on top of representations, we added implementations of Random Forest (RF) and Naive Bayes (NB) from scikit-learn as other popular but `simple' classifiers. SentEval defines specific model validation techniques for each task. Following SentEval,  
we tune the size of the hidden layer in $\{50, 100, 200\}$, dropout in $\{0.0, 0.1, 0.2\}$ and $L^2$ regularization in $\{10^{-5}, 10^{-1}\}$ when training an MLP.
For RF, we tune the maximum tree depth in $\{10, 50, 100, \infty\}$.
For LR, we tune the $L^2$ regularization in $\{10^{-5}, 10^{-1}\}$. We do not tune any hyperparameters for NB. 
For all probing tasks and TREC, we use predefined splits. 
For AM and sentiment analysis, we use 10-fold inner cross validation.

\subsection{Probing task design in \en{}}
\label{sec:Dimensionalities-of-probing-task-design}
In our design, we consider (a) four  well-known and popular classifiers---LR, MLP, NB, RF---on top of sentence representations,  and (b) six different training data sizes (between 2k and 100k). 
We perform an exhaustive grid-search for size and classifier design, considering all combinations. 

\paragraph{Size} For each classifier, we obtain results (on 10k test instances) when varying the training data size 
over 2k, 5k, 10k, 20k, 30k, 100k. 

Downsampling was implemented by selecting the same percentage of samples that appears in the full dataset for each class. 
We then report average Spearman/Pearson correlations $\rho$/$p$  between any two training set sizes $s$ and $t$ over all 9 probing tasks:\footnote{We report both Spearman and Pearson for some of the results but give only Pearson for the remainder, 
reporting Spearman in the appendix.}
\begin{align}\label{eq:1}
    \text{sim}_c(s,t)=\frac{1}{n}\sum_{i=1}^{n} \rho^* (\textbf{c}_i(s),\textbf{c}_i(t))
\end{align}
where $n$ is 
the number of 
probing tasks ($n=9$ for \en{}), and $\textbf{c}_i(s)$ is the vector that holds scores for each of the 7 sentence encoders in our experiments, given training size $s$, for probing task $i$ and classifier $c$. We set 
correlations to zero if the p-value $>0.2$.\footnote{We choose a high p-value, because we correlate small vectors of size 7.} In Table \ref{table:sizes}, we then report the minimum and average scores  $\min_{(s,t)}\text{sim}_c(s,t)$ and $\frac{1}{M}\sum_{(s,t)}\text{sim}_c(s,t)$, respectively, per classifier $c$. 
We observe that the minimum values are small to moderate correlations between 0.2 (for NB) and 0.6 (for RF). The average correlations are moderate to high correlations ranging from 0.6 (for NB) to above 0.8 (for the others). 

\begin{figure*}
    \centering
    \includegraphics[scale=0.1025]{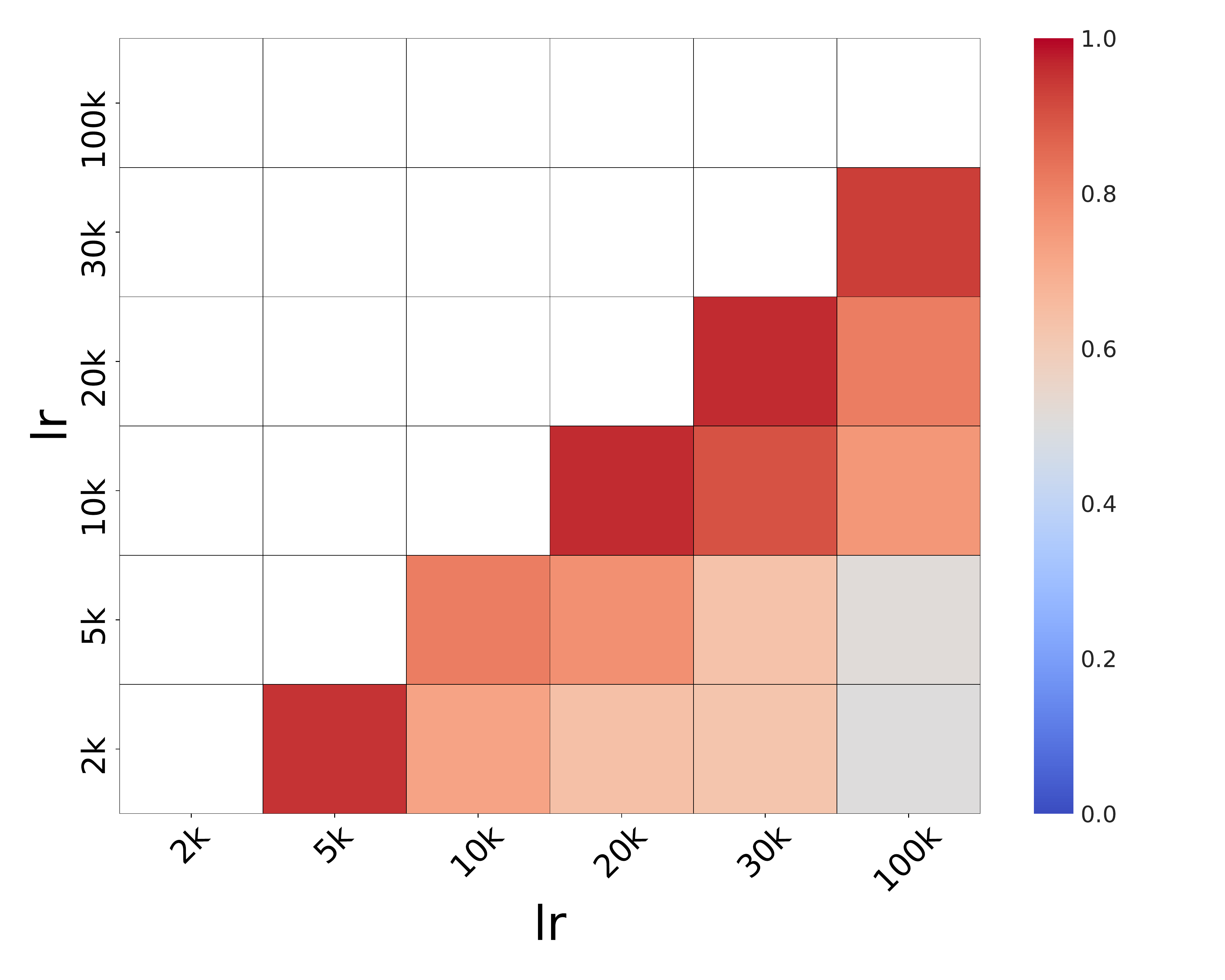}
    \includegraphics[scale=0.1025]{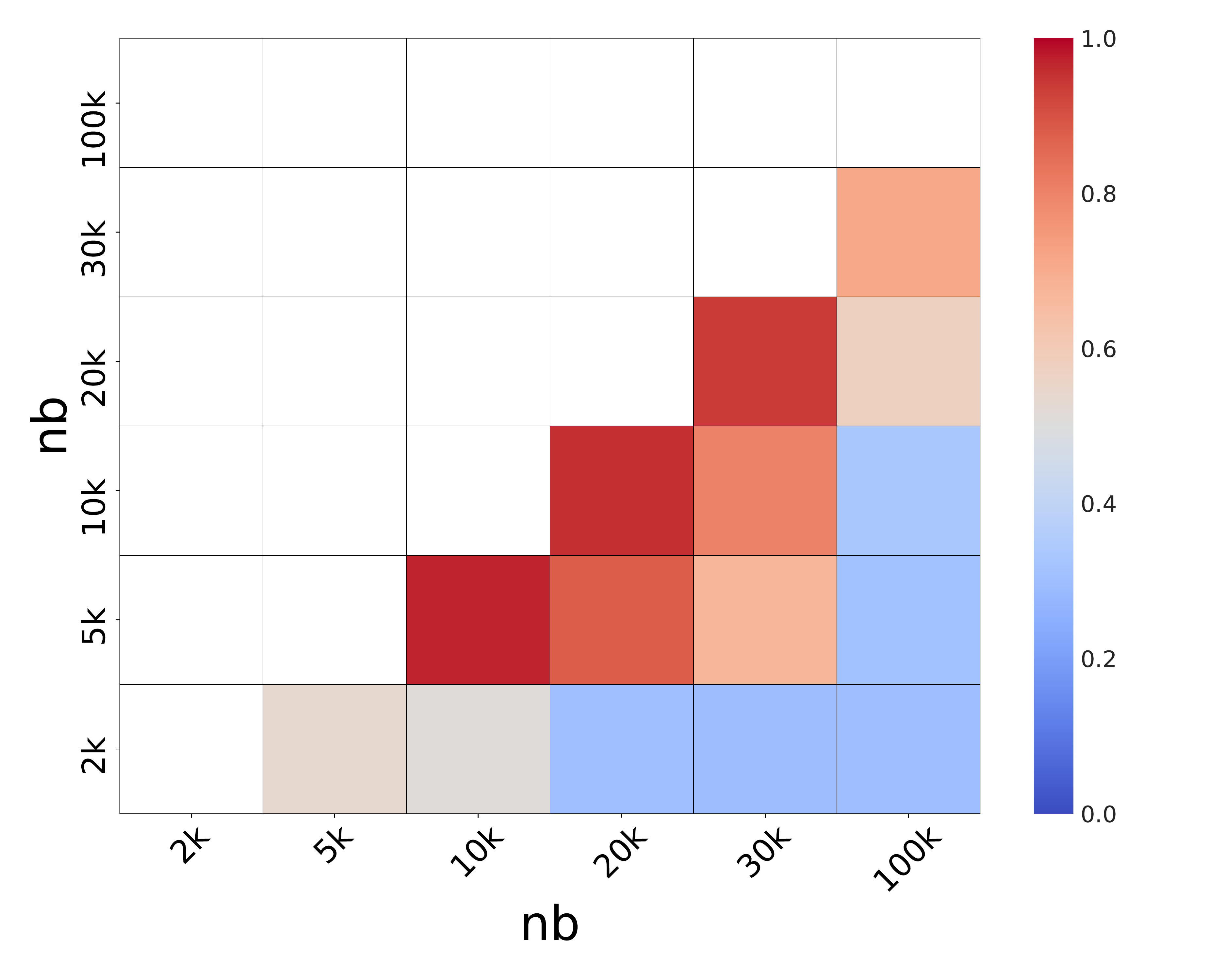}
    \includegraphics[scale=0.1025]{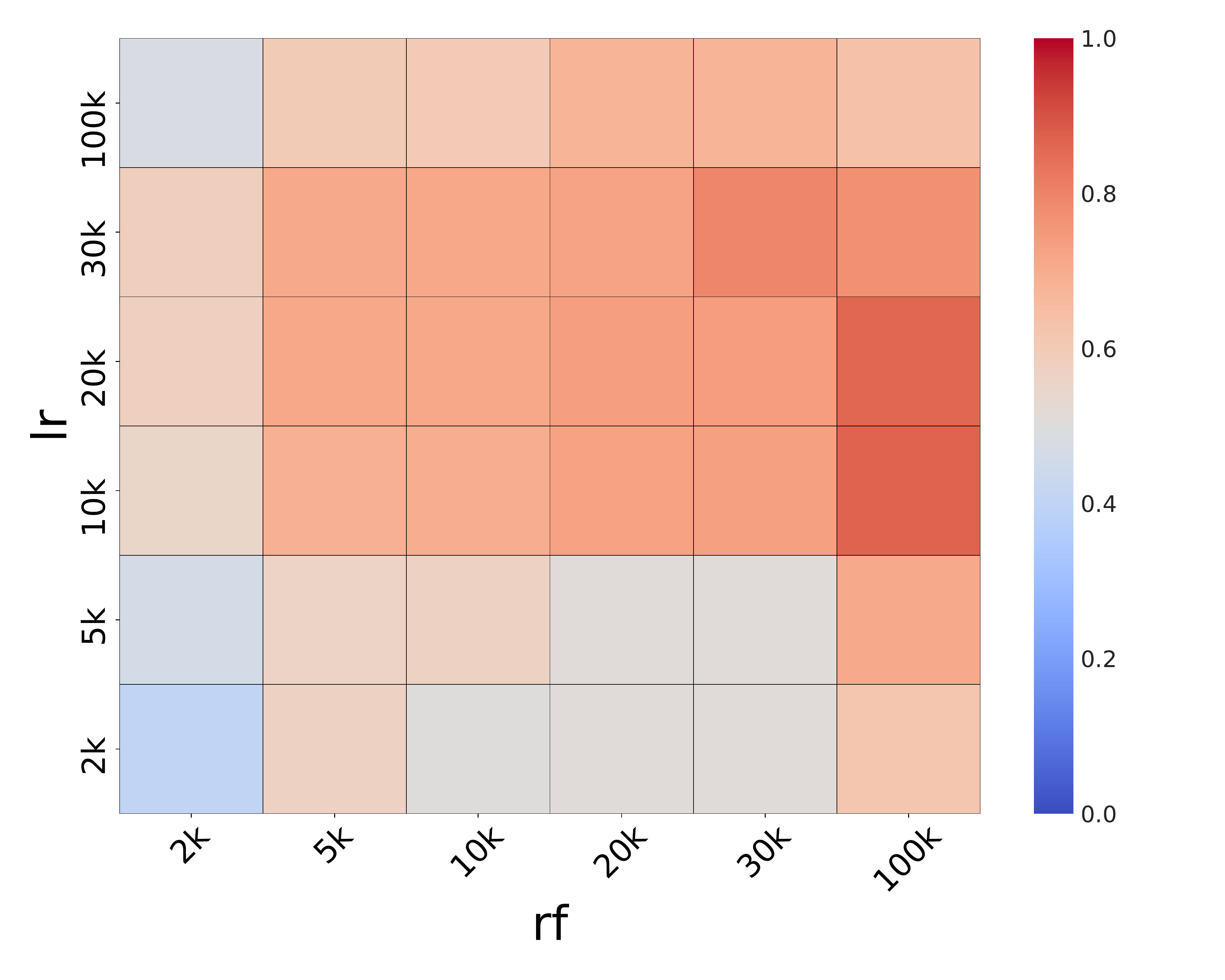}
    \includegraphics[scale=0.1025]{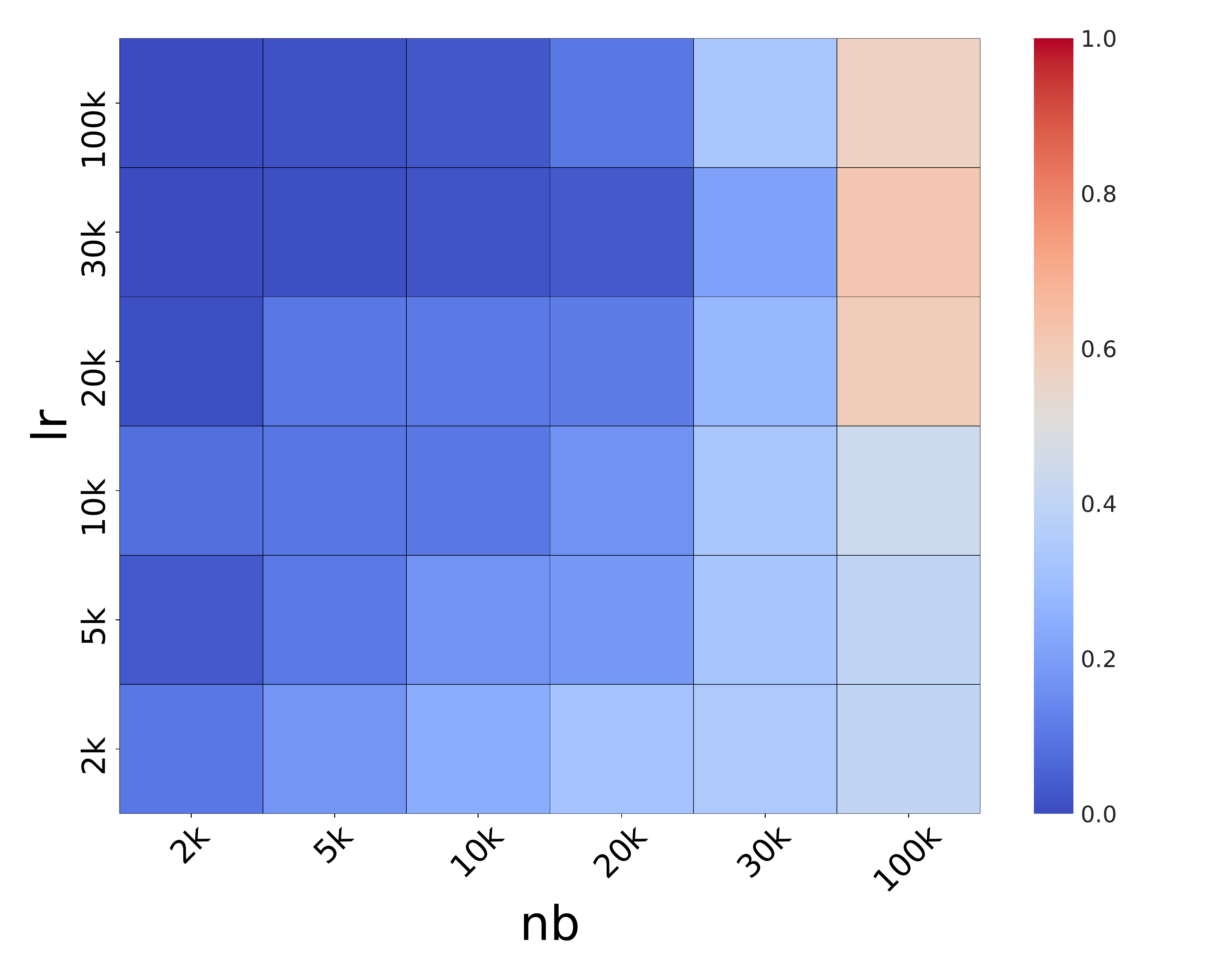}
    \caption{Left two: Average correlations sim$_c(s,t)$ for LR (first) and NB (second), using Pearson, for any two sizes $s,t\in\{2k,5k,10k,20k,30k,100k\}$. Right two: Average correlations sim$_{c,d}(s,t)$ for $c=$ LR and $d=$ RF (third) and $c=$ LR and $d=$ NB (fourth). \se{Best viewed in color}.}
    \label{figure:nb_lr}
\end{figure*}

In Figure \ref{figure:nb_lr} (left), we show  
all the values sim$_c(s,t)$ for $c=$ LR, NB. 
We observe that, indeed, LR has  
high correlations between training sizes especially starting from 
10k training data points. 
The corresponding correlations of NB are much lower comparatively.

\begin{figure}
  \centering
  \includegraphics[scale=0.45]{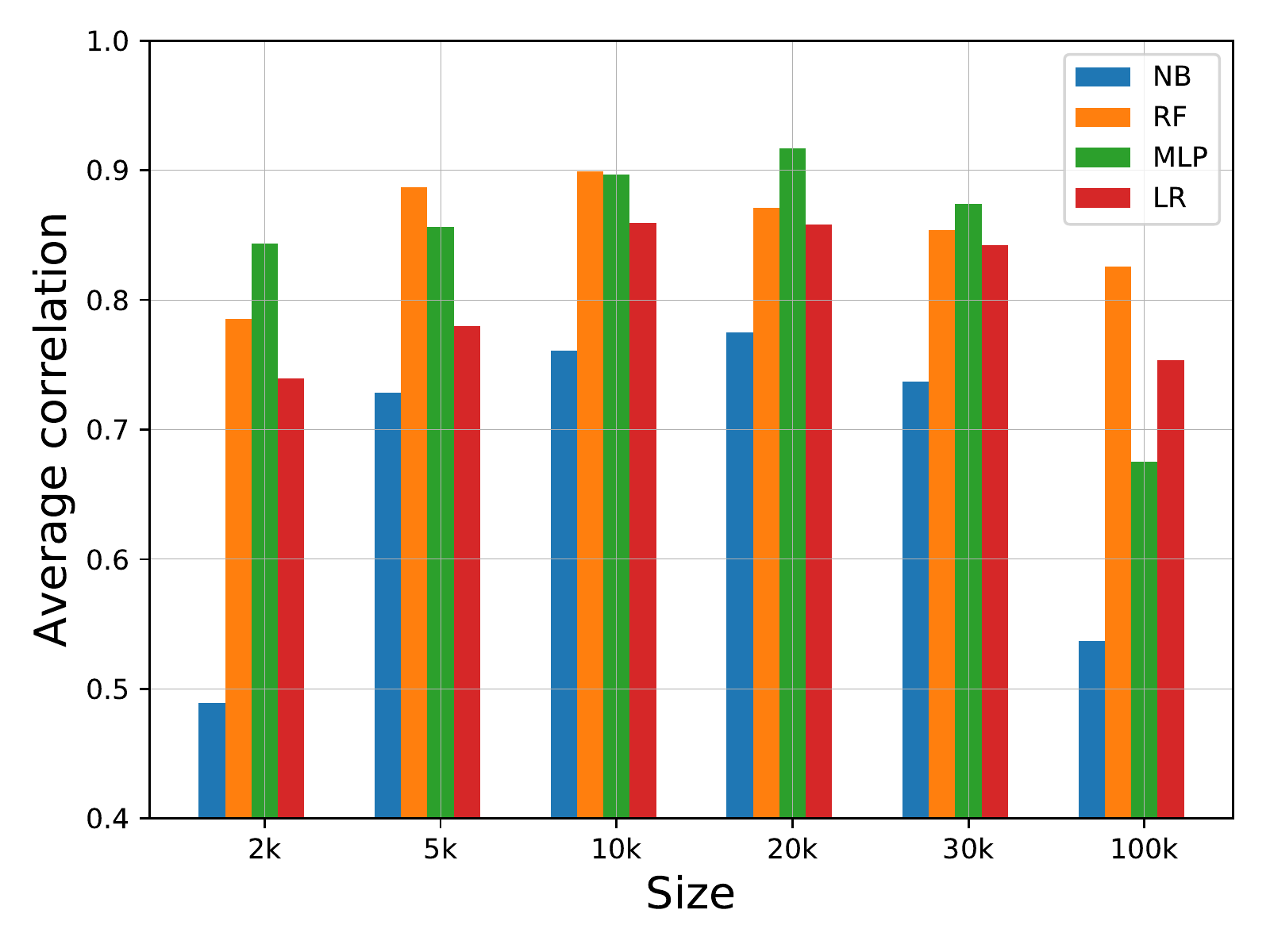}
  \caption{Stability of each training size computed using Eq.~\eqref{eq:size_stability} for different classifiers $c$.}
  \label{figure:intraclassifier}
\end{figure}

In Figure \ref{figure:intraclassifier}, we plot the stability of each training data size $s$ 
\begin{align}\label{eq:size_stability}
    \text{sim}_c(s) = \frac{1}{N}\sum_{t}\text{sim}_c(s,t)  
\end{align}
for all of our classifiers $c$ and where $N$ is a normalizer equal to the number of different training set sizes, $N=6$ in our case. 
The higher this score for a training size $s$, the more similar are the probing results for another training size $t$, on average. 
Across all classifiers, 2k and 100k are least stable---100k is the default setting of SentEval. Most stable are 10k and 20k. 

\begin{table}[!htb]
    \centering
    {\footnotesize
    \begin{tabular}{l|rr|rr} \toprule
          & \multicolumn{2}{c|}{\textbf{Min}} & \multicolumn{2}{c}{\textbf{Avg}} \\
         \textbf{Classifier}& $\rho$ & $p$ & $\rho$ & $p$\\
         \midrule 
         MLP & .480 & .420 &.810 & .843\\
         LR & .524 & .502 & .808 & .805\\
         RF & .529 & .623 & .800 & .853\\
         NB & .174 & .292 & .626 & .671\\
         \bottomrule
    \end{tabular}
    }
    \caption{Stability over training sizes, in terms of minimum and average Spearman ($\rho$) / Pearson ($p$) correlation between any two sizes.}
    \label{table:sizes}
\end{table}

\paragraph{Classifier}  Next, we 
add the classifier choice as a second dimension: 
we examine whether correlations (Spearman/Pearson) between vectors $\mathbf{c}$ (holding scores for each of 7 sentence encoders for a classifier $c$) and $\mathbf{d}$ (holding the same scores for a classifier $d$) are similar in the same sense as in Eq.~\eqref{eq:1}: 
\begin{align}\label{eq:2}
    \text{sim}_{c,d}(s,t)=\frac{1}{n}\sum_{i=1}^{n} \rho^* (\textbf{c}_i(s),\textbf{d}_i(t))
\end{align}
Again, we average across all probing tasks, and set correlation values to zero if the p-value exceeds 0.2. 
In Table \ref{table:cross_correlations_classifiers}, we give min/avg values across data set sizes in this setup. We observe that LR and MLP most strongly agree. 
They have acceptable average agreement with RF, but low agreement with NB, on average, and, in the worst cases, 
even negative correlations with NB. 

\begin{table}[!htb]
    \centering
    { \small
    \begin{tabular}{l|cccc}
    \toprule
             & LR & RF & NB \\ \midrule
         MLP &  .481/.790 & .492/.632& -.043/.236 \\
         LR &  & .406/.640 & -.057/.197\\
         RF &  & & .029/.320\\
         \bottomrule
    \end{tabular}
    }
    \caption{Min/Avg values sim$_{c,d}(s,t)$ across $(s,t)$ (using Pearson) between classifiers $c$ and $d$.}
    \label{table:cross_correlations_classifiers}
\end{table}

In Figure \ref{figure:nb_lr} (right), 
we illustrate correlations between 
LR and NB, on the one side, 
and LR and RF, on the other side, 
across all possible training set sizes. We observe that as the training data set sizes for RF and LR become larger, these two classifiers agree more strongly with LR. 
RF starts to have acceptable agreement with LR from 10k training instances onwards, while NB has acceptable agreement with LR only in the case of 100k training instances. 

We now operationalize our intuition of `region of stability' outlined in Table \ref{table:stability}. For each of nine probing tasks, we 
compute the following. 
Let $r_j = E_{\zeta(1)}\succ E_{\zeta(2)}\succ E_{\zeta{(3)}} \succ \cdots$ be a specific ranking of encoders, where $\zeta$ is a fixed permutation. 
Let $r_{(c,s)}$ be the ranking of encoders according to the classifier, size combination $(c,s)$. We compute the Spearman correlation $\tau_{(c,s,j)}$ between $r_{(c,s)}$ and $r_j$. 
For each possible ranking $r_j$ of our 7 encoders, we then determine its support as the average over all values $\tau_{(c,s,j)}$ and then find the ranking $r_{\text{max}}$ with most support according to this definition. Finally, we assign a score to the combination $(c,s)$ 
not only when $r_{(c,s)}$ equals $r_{\text{max}}$, but also when $r_{(c,s)}$ is close to $r_{\text{max}}$: we again use the Spearman correlation between $r_{(c,s)}$ and $r_{\text{max}}$ as a measure of closeness (we require a closeness of at least 0.75). The final score for $(c,s)$ is given by (\se{ignoring the threshold of 0.75 in the equation}):
\begin{align}\label{eq:stabl}
    \mu_{(c,s)} = \sum_{i=1}^{n} \rho^*(r^{(i)}_{(c,s)},r^{(i)}_{\text{max}})
\end{align}
Table \ref{table:stability_comb} shows 
classifier, size combinations with highest $\mu$ scores. LR and MLP are at the top, along with RF in the setting of 100k training data size. LR with size 10k is most stable overall, but the distance to the other top settings is small. Least stable (not shown) is NB.  
\begin{table}[!htb]
    \centering
    {\small
    \begin{tabular}{l|cccccc}
    \toprule
        \textbf{classifier} & LR & LR & RF & MLP & MLP & MLP\\
        \textbf{size} & 10k & 20k & 100k & 20k & 30k & 10k \\ \midrule
        \textbf{$\mu_{(c,s)}$} & 7.6 & 7.3 & 7.2 & 7.0 & 7.0 & 6.9\\ 
         %
         \bottomrule
    \end{tabular}
    }
    \caption{Most stable classifier, size combinations according to Eq.~\eqref{eq:stabl}.}
    \label{table:stability_comb}
\end{table}

\se{Overall, we answer our  
first research question---\textbf{(i) How reliable are probing task results across
machine learning design choices?}---as follows (for \en{})}: probing tasks results can be little reliable and may vary greatly among machine learning parameter choices. The standard training set size of SentEval, 100k, appears to be less stable. As 
region of stability, 
we postulate especially the setting with 10k training instances for the LR classifier. 

\subsection{Multi-lingual results}
\paragraph{Experimental Setup}
Given our results for \en{}, we choose the LR classifier with a size of roughly 10k instances overall.  
Table \ref{table:multilingual_probing} provides more details about the datasets. In line with SentEval (and \se{partly} supported by our results on dataset balance given in the appendix), we aim for as balanced label distributions as possible.  
Because of the small test sizes, we use inner 5-fold cross validation for all 
tasks except for SubjNumber, where we use pre-defined train/dev/test splits as in \citet{Conneau.2018a} to avoid leaking lexical information from train to test splits.

We obtain average and pmeans embeddings through pooling over pre-trained FastText embeddings \citep{grave2018learning}. The same embeddings are used for the random LSTM.
For average BERT, we use the base-multilingual-cased model.
We machine translate the AllNLI corpus into \tr{}, \ru{} and \ka{}, to obtain training data for Infersent.\footnote{Using Google Translate, see appendix for details.} 
The models are then trained using default hyperparameters and with  pre-trained FastText embeddings.
Compared to \en{}, we 
modify the WC probing task in the multilingual setting to only predict 30 mid-frequency words instead of 1000. This is more appropriate for our much smaller data sizes. 
\begin{table*}[!htb]
    \centering
    {\footnotesize
    \begin{tabular}{l|rr|rr|rr|rr} \toprule
         & \multicolumn{2}{c|}{\textbf{EN}} &
         \multicolumn{2}{c|}{\textbf{TR}} & \multicolumn{2}{c|}{\textbf{RU}} & \multicolumn{2}{c}{\textbf{KA}} \\
         \textbf{Task} & \textbf{Size} & \textbf{Balance} &\textbf{Size} & \textbf{Balance} & \textbf{Size} & \textbf{Balance} & \textbf{Size} & \textbf{Balance} \\
         \midrule 
         Bigram Shift & 100k & 1:1 & 10k & 1:1 & 10k & 1.1:1 & 10k & 1.1:1 \\
        Length & 100k & 1:1 & 10k & 1:1 & 12k & 1:1 & 10k & 1:1 \\
         Subject Number & 100k & 1:1 & 4,093 & 5:1 & 11,877 & 1:1 & - & - \\
         Word Content & 100k & 1:1 & 10k & 1.5:1 & 10k & 1.2:1 & 10k & 4:1 \\
         Top Constituents & 100k & 1:1 & - & - & - & - & - & - \\
         Tree Depth & 100k & 2.2:1 & - & - & - & - & - & - \\
                  Voice & 100k & 1:1 & 8,417 & 6:1 & 10k & 2:1 & 10k & 1.9:1 \\
SV-Agree & 100k & 1:1 & 10k & 1:1 & 10k & 1:1 & 10k & 1:1 \\
         SV-Dist & 100k & 1:1 & 1,642 & 1.9:1 & 8,231 & 1.1:1 & - & - \\
         \midrule
         Arg. Mining (macro-F1) & 25,303 & 3:1 & 25,303 & 3:1 & 25,303 & 3:1 & 25,303 & 3:1 \\
         TREC (Accuracy) & 5,952 & 14:1 & 5,952 & 14:1 & 5,952 & 14:1 & 5,952 & 14:1 \\
         Sentiment Analysis (macro-F1) & 14,148 & 4.2:1 & 6,172 & 1.7:1 & 30k & 1:1 & 11,513 & 5.5:1 \\
         \bottomrule
    \end{tabular}
    }
    \caption{Probing and downstream tasks. We report the balance between the class with the most and the least samples. For downstream tasks, the evaluation measure is given in brackets.}
    \label{table:multilingual_probing}
\end{table*}

\subsubsection{Probing tasks}
\begin{figure}[!htb]
    \centering
    \includegraphics[scale=0.15]{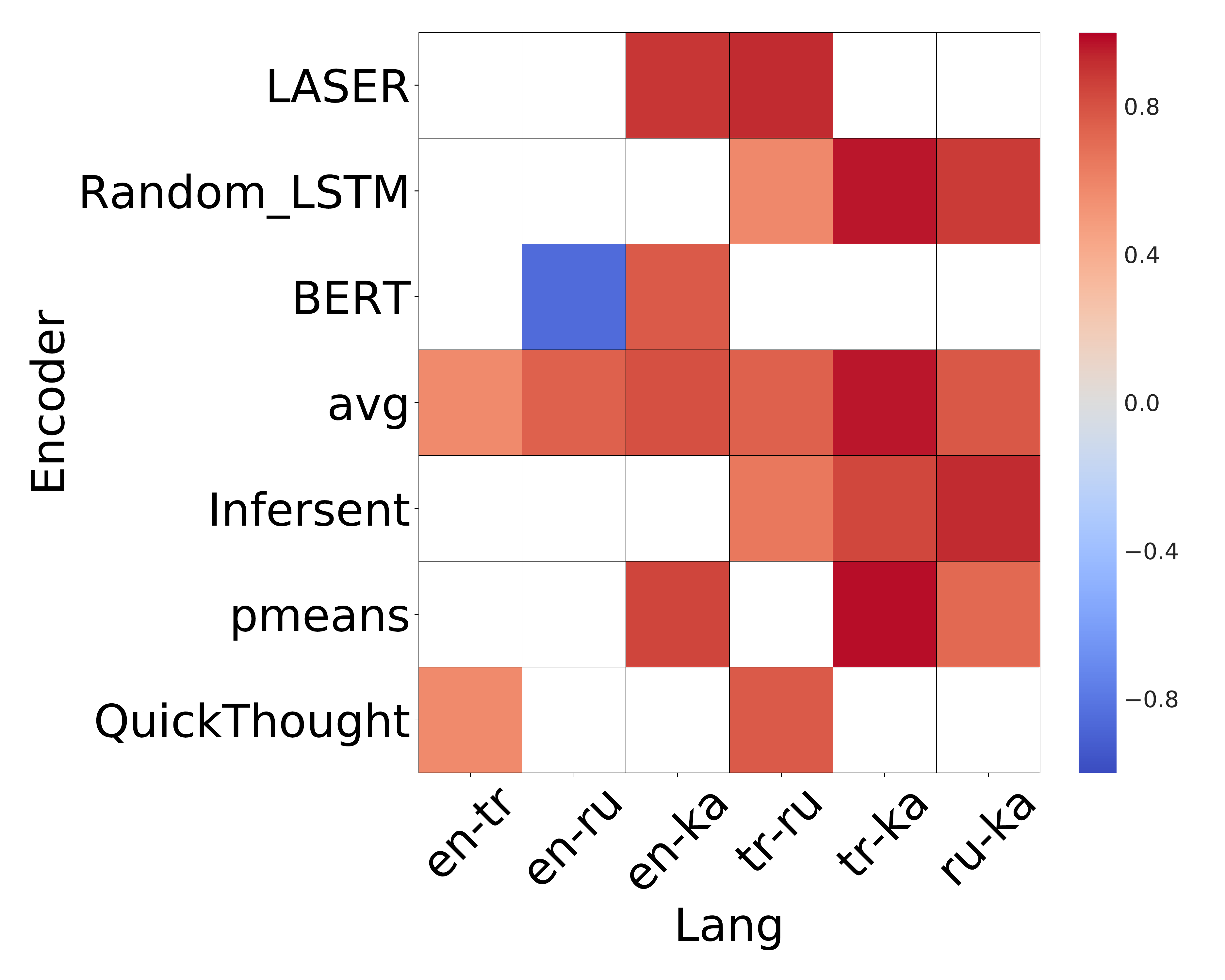}
    \caption{Pearson correlations 
    across languages for different encoders.}
    \label{figure:encoder_lang}
\end{figure}

\begin{figure}[!htb]
    \centering
    \includegraphics[scale=0.15]{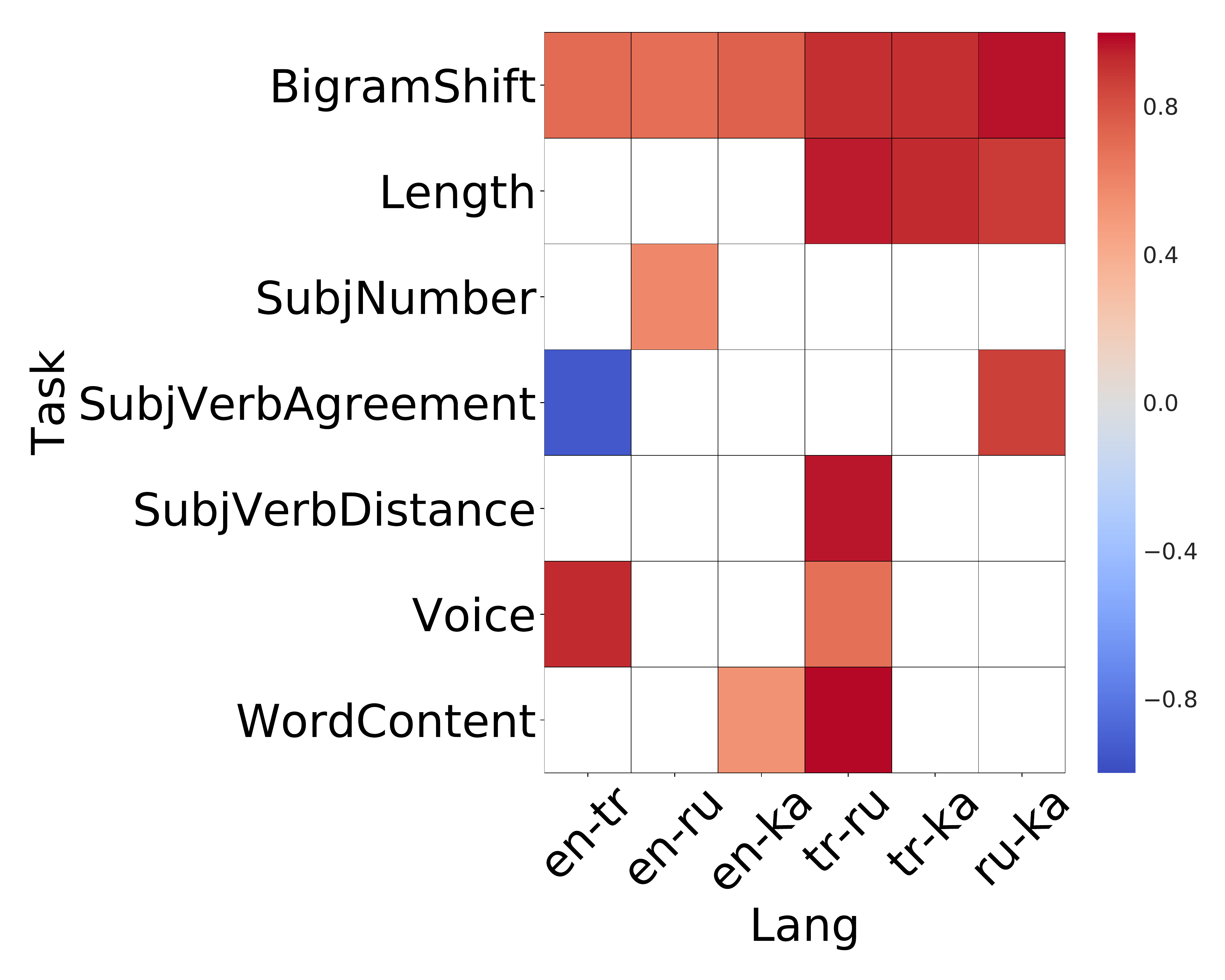}
    \caption{Pearson correlations 
    across languages for different probing tasks.}
    \label{figure:task_lang}
\end{figure}

Results are shown in Figures \ref{figure:encoder_lang} and \ref{figure:task_lang}.\todo{SE: do we always report micro-F1 as in SentEval because datasets are mostly balanced?} 

\paragraph{(ii) Will encoder performances correlate across languages?}
For each encoder $e$, we correlate performances of $e$ between 
\en{} and the other languages on 5 (for \ka{}) and 7 (for \tr{}, \ru{}) probing tasks (using 10k dataset size and LR for all involved languages, including \en{}). 
In Figure \ref{figure:encoder_lang}, we observe that correlations between \en{} and other languages are generally either zero or weakly positive. 
Only average embeddings have more than 1 positive correlation scores across the 3 language combinations with \en{}. Among low-resource languages, there are no negative correlations and fewer zero correlations. All of the low-resource languages correlate more among themselves than with \en{}. This makes sense from a linguistic point of view, since \en{} is clearly the outlier in our sample given its minimal inflection and fixed word order. Thus, the answer to this research question is that our results support the view that transfer is 
better for typologically similar languages. 

\paragraph{(iii) Will probing task performances correlate across languages?} 
For each probing task $\pi$, we report 
Pearson correlations, between all language pairs, of vectors holding scores of 7 encoders on $\pi$. 
Figure \ref{figure:task_lang} shows the results. 
The pattern is overall similar as for (ii) in that there are many zero correlations between \en{} and the other languages. \tr{} also negatively correlates with \en{} for SV-Agree. Only BigramShift has positive correlations throughout. 
Low-resource languages correlate better among themselves as with \en{}. 
Our conclusions are the same as for question \textbf{(ii)}. 

Note that our findings contrast with 
\newcite{Krasnowska.2019}, who report that probing results for \en{} and \pl{} are mostly the same. 
Our results are more intuitively plausible: 
e.g., a good encoder should store linguistic information relevant for a particular language. 

\begin{figure*}[!htb]
    \centering
    \includegraphics[scale=0.1025]{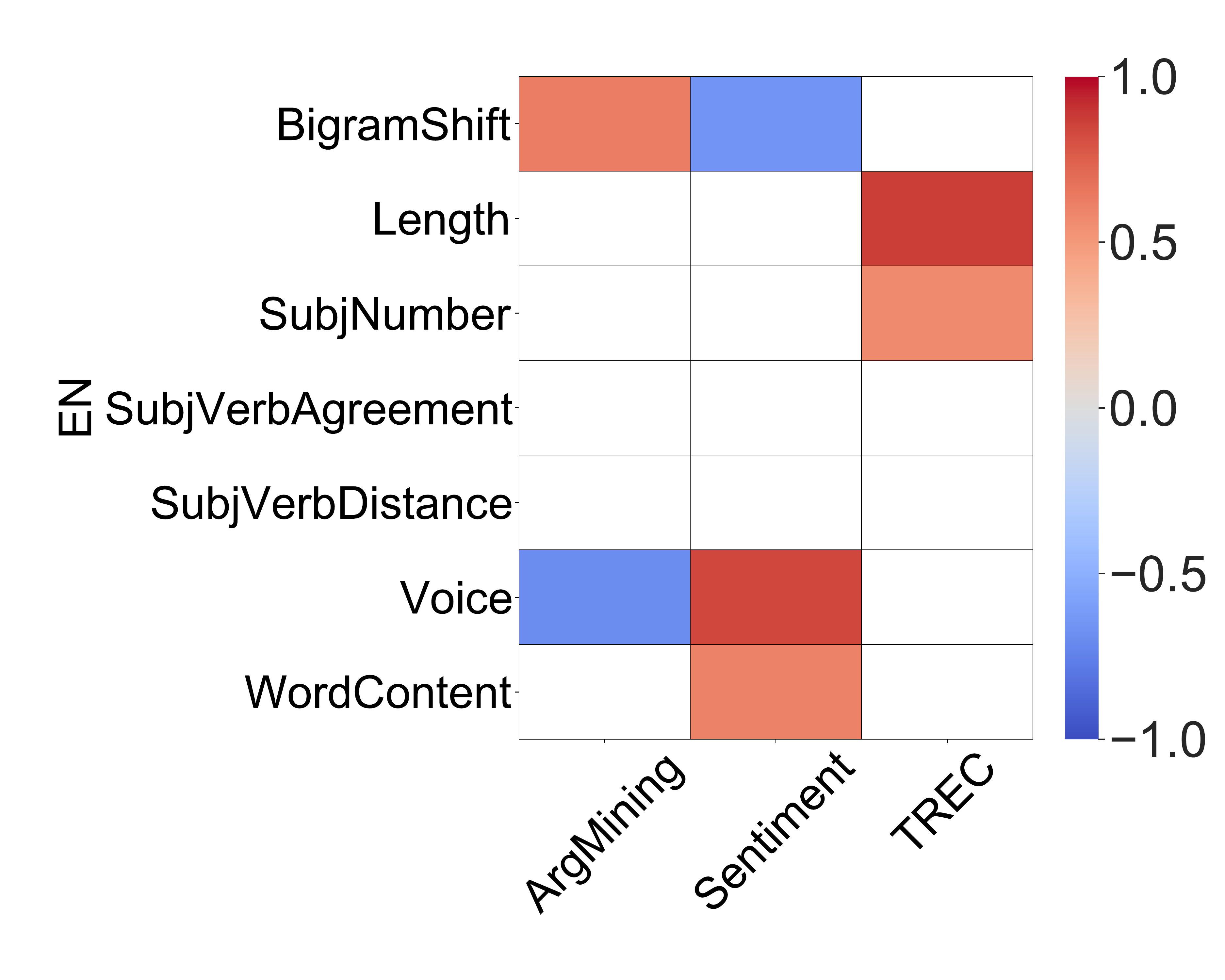}
    \includegraphics[scale=0.1025]{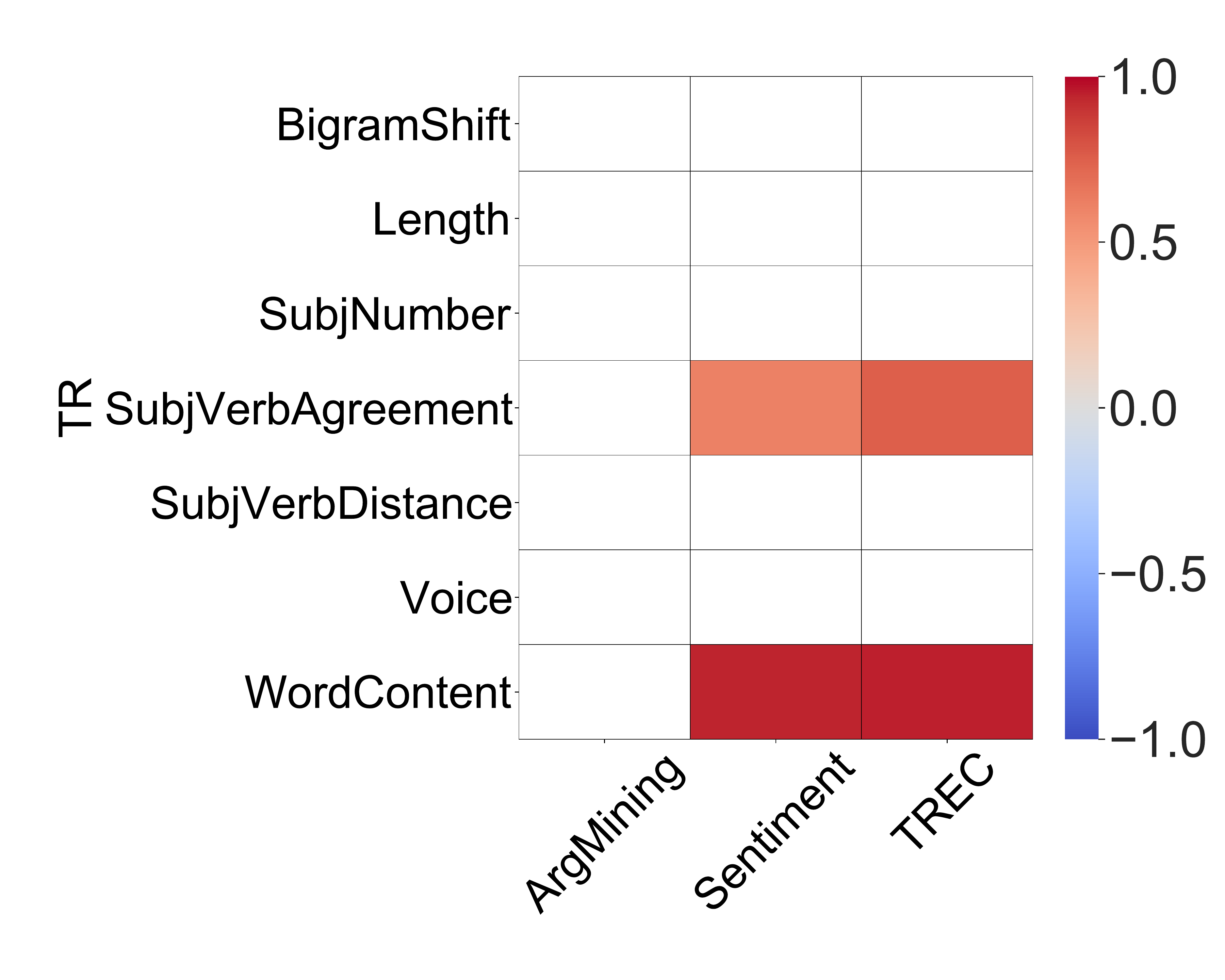}
    \includegraphics[scale=0.1025]{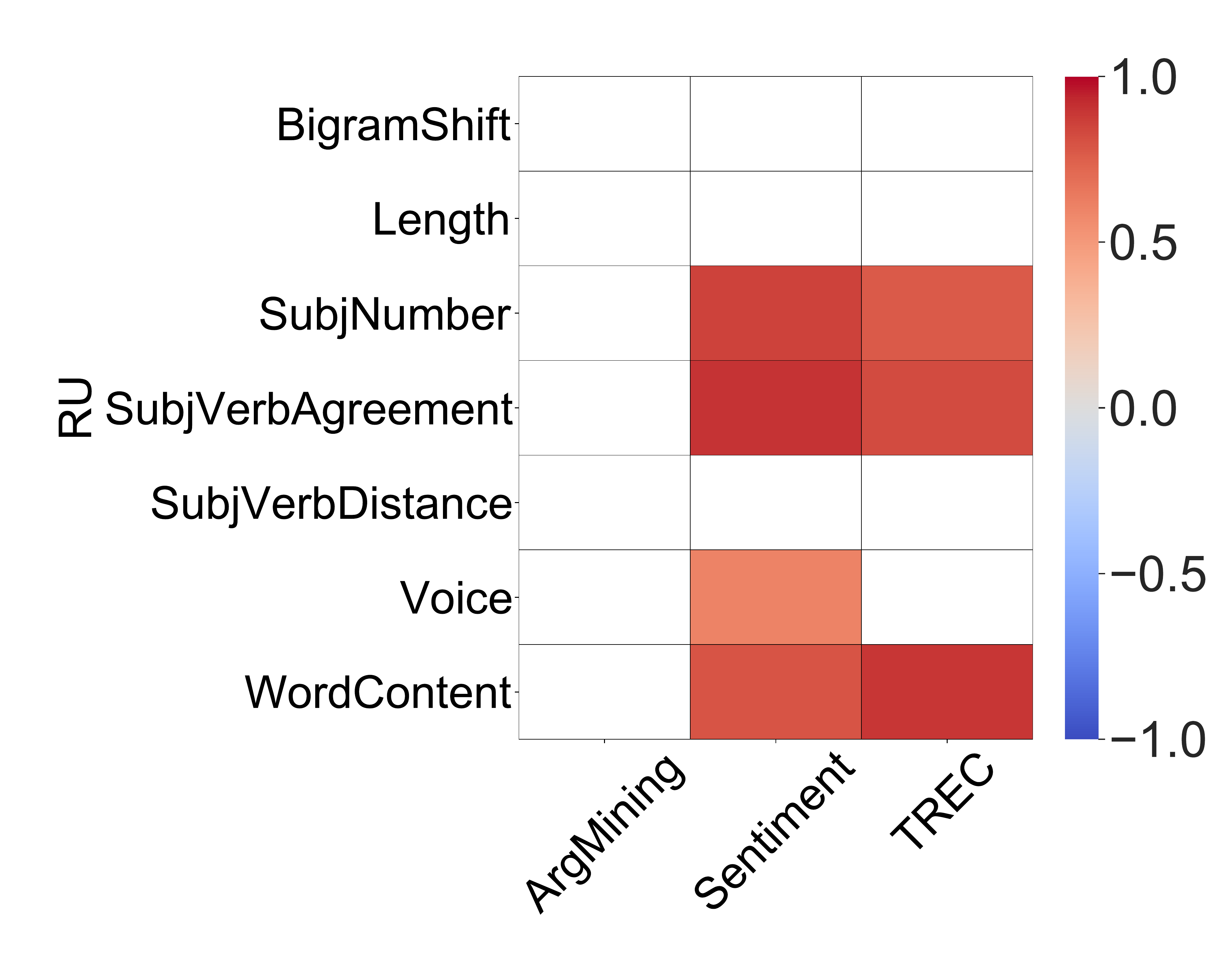}
    \includegraphics[scale=0.1025]{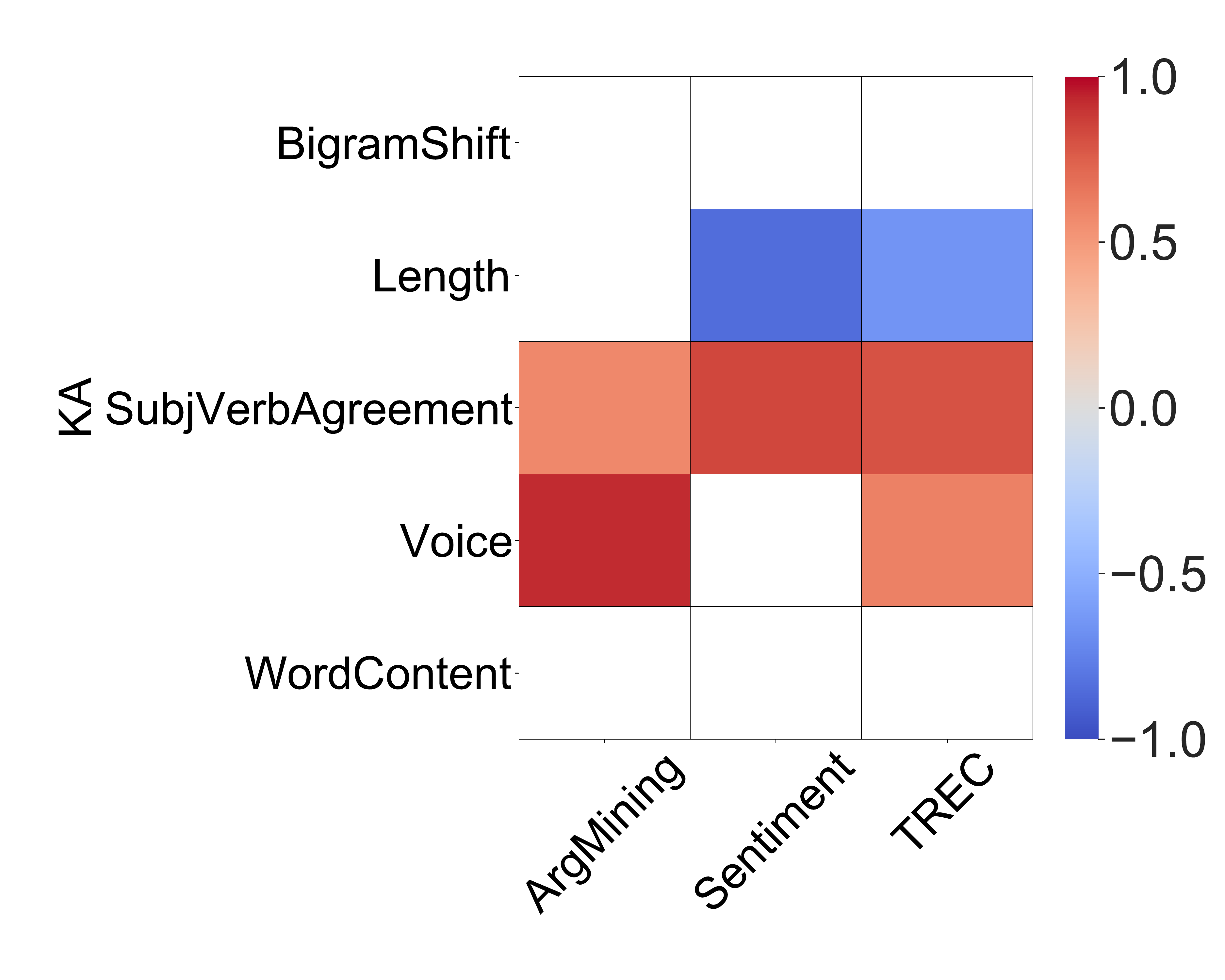}
    \caption{Pearson correlation among probing task and downstream performance for all languages.}
    \label{figure:downstream_multi}
\end{figure*}

\subsubsection{Downstream Tasks}
Results are shown in Figure \ref{figure:downstream_multi}.

\paragraph{(iv) Will the correlation between probing and downstream tasks be the same across languages?} 

For each of our languages, we correlate 
probing  
and downstream task performances.
The results 
show that the answer to research question (iv) is clearly negative. 
In particular, \en{} behaves differently to the other languages---while \ru{} and \tr{} 
behave more similarly. 
\ka{} is the only language with negative correlations for Length, \en{} the only one with positive scores.
For the sentiment task, Word Content correlates positively for all languages except \ka{}.
The AM task correlates only in \en{} and \ka{}, 
but with different probing tasks. 
SV-Agree correlates positively with TREC and sentiment in all languages but \en{}.
\se{This might be because determining the agreement of subject and verb is more grammatically complex in the other languages compared to English, and storing an adequate amount of grammatical information may be beneficial for certain downstream tasks.}  
Predicting the performances of embeddings in downstream tasks via probing tasks thus appears idiosyncratic for individual languages. Opposed to \newcite{DBLP:journals/corr/abs-1903-09442}, who suggest a direct relation between word level probing and downstream performance on agglutinative languages, we see little to no systematic correlation on the sentence level.
Overall,  
SV-Agree 
is the best predictor across languages, with 7 positive correlations out of 12 possible. Interestingly, this task is missing from the current canon of SentEval. 
\section{Concluding Remarks}
We investigated formal aspects of probing task design, including probing data size and classifier choice, in order to determine structural conditions for multilingual (low-resource) probing. 
We showed that probing tasks results are at best partly stable even for \en{} and that the rankings of encoders 
varies with design choices. 
However, we identified a partial region of stability where results are supported by a majority of  settings---even though this may not be mistaken for a `region of truth'. This region was identified in \en{}, which has most resources available. Our further findings then showed that probing and downstream results do not transfer well from English to our other languages, which in turn challenges our identified region of stability. 

Overall, our results 
have partly negative implications for current practices of probing task design 
as they indicate  
that probing tasks are to some degree unreliable 
tools for introspecting linguistic information contained in sentence encoders. 
The relation of probing to downstream tasks is also unclear, as our multilingual results show.  
This is supported by recent findings giving contradictory claims regarding, e.g., the importance of the Word Content probing task for downstream performances \citep{Eger.2019,wang2020sbertwk,Perone.2018}.
Our findings further add to contemporaneous work by \newcite{Ravichander2020ProbingTP} and \newcite{Elazar2020WhenBF}, who showed that probes do not necessarily identify linguistic properties 
required for solving an actual task, 
thus questioning a common interpretation of probing itself.

An important aspect to keep in mind 
for correlation analyses as we conducted  
is that results may heavily depend on the selection of encoders involved---
in our case, we selected a number of recently proposed state-of-the-art models in conjunction with weaker baseline models, for a diverse collection of encoders. 
While the small number of encoders we examined is a clear limitation of our approach, many of our results are significant (at relatively large p-values). 

To the degree that the supervised probing tasks examined here will remain important tools for interpretation of sentence encoders in the future, our results indicate that multilingual probing is important for a fairer and more comprehensive comparison of encoders.

\section*{Acknowledgments}
We thank the anonymous reviewers for their useful comments and suggestions. 
Wei Zhao and Benjamin Schiller gave useful feedback on earlier versions of this paper. 
Daniel Wehner and Martin Kerscher greatly helped to extend the SentEval code and collect the probing data for the low-resource languages.
We finally thank Paul Meurer from the University of Bergen for making the Georgian National Corpus (GNC) available for our experiments in \ka.
The first author has been funded by the HMWK (Hessisches Ministerium für Wissenschaft und Kunst) as part of structural location promotion for TU Darmstadt in the context of the Hessian excellence cluster initiative ``Content Analytics for the Social Good'' (CA-SG).  
The second author has been supported by the German Federal Ministry of Education and Research (BMBF) under the promotional reference 03VP02540 (ArgumenText).

\bibliography{bibliography2020}
\bibliographystyle{acl_natbib}
\clearpage
\appendix

\section{Quality of Machine Translated Data}

\begin{table}[!htb]
    \centering
    { \small
    \begin{tabular}{l|ccc}
    \toprule
             & BLEU (1-gram) & METEOR & MoverScore 
              \\ \midrule
         en-ka & 0.271 & 0.149 & 0.272 \\
         en-ru & 0.470 & 0.335 & 0.353 \\
         en-tr & 0.493 & 0.359 & 0.398 \\ 
         \midrule
         en-de & 0.446 & 0.342 & 0.337 \\
         \bottomrule
    \end{tabular}
    }
    \caption{Quality of the machine translation service used to translate training data for downstream tasks on reference datasets.}
    \label{table:mt-scores}
\end{table}

We automatically translated the input data for the AM and TREC downstream tasks.
To estimate the quality of the machine translated data, we measured the performance of the service used to translate the data with the help of the JW300 corpus \cite{agic-vulic-2019-jw300,TIEDEMANN12.463}.
For each of the language pairs \en{}-\ka{}, \en{}-\tr{}, and \en{}-\ru{}, we translated the first 10,000 sentences of the respective bitext files from JW300 and measured their quality in terms of BLEU, METEOR and MoverScore  \citep{zhao-etal-2019-moverscore}.\footnote{Misaligned sentences were skipped.} 
As reference, we also added \en{}-\de{} as a pair with two well-resourced languages.
Results are summarized in Table~\ref{table:mt-scores}.
They show that, with the exception of \en{}-\ka{}, all language pairs have high-quality translations (on par or even better than \en{}-\de{}).
We thus expect the influence of errors of the machine translated data to be minimal in 
\tr{} and \ru{}. 
For \ka, this is not necessarily the case.

\section{Sentence Encoder Dimensions} 
Table~\ref{table:encoders} shows the full list of encoders used in our study and their dimensionalities.
\begin{table}[!htb]
    \centering
    {\footnotesize
    \begin{tabular}{l|rrr} \toprule
         \textbf{Encoder} & \textbf{Size} & 
         \\
         \midrule 
         Avg & 300 \\
         pmeans (Avg$+$Max$+$Min) & 900 \\
         Random LSTM & 4096\\
         \midrule
         InferSent & 4096 \\
         QuickThought & 2400 \\ 
         LASER & 1024 \\
         BERT & 768 \\
         \bottomrule
    \end{tabular}
    }
    \caption{Encoders and their dimensionalities.} 
    \label{table:encoders}
\end{table}

\begin{figure}[!htb]
    \centering
    \includegraphics[scale=0.09]{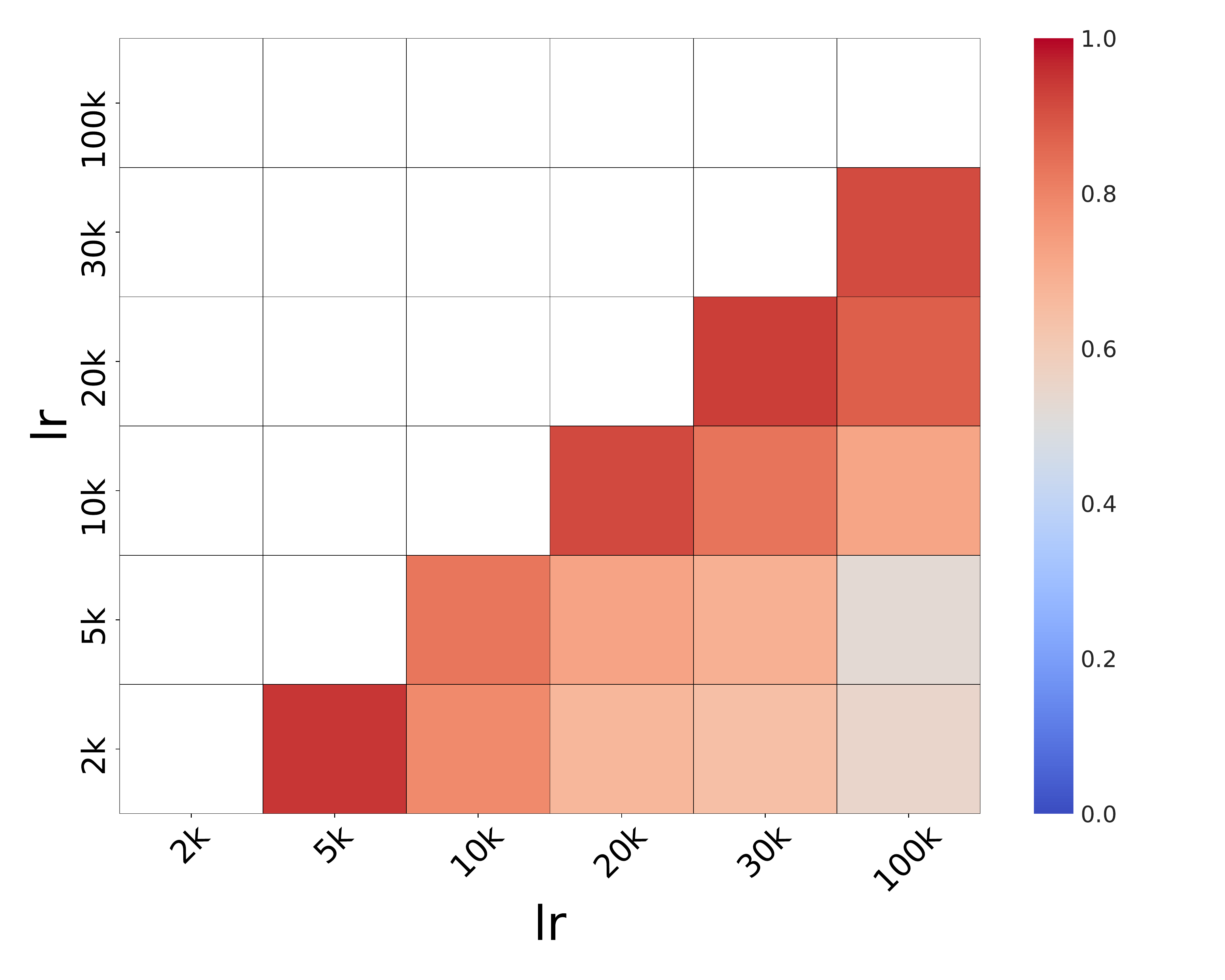}
    \includegraphics[scale=0.09]{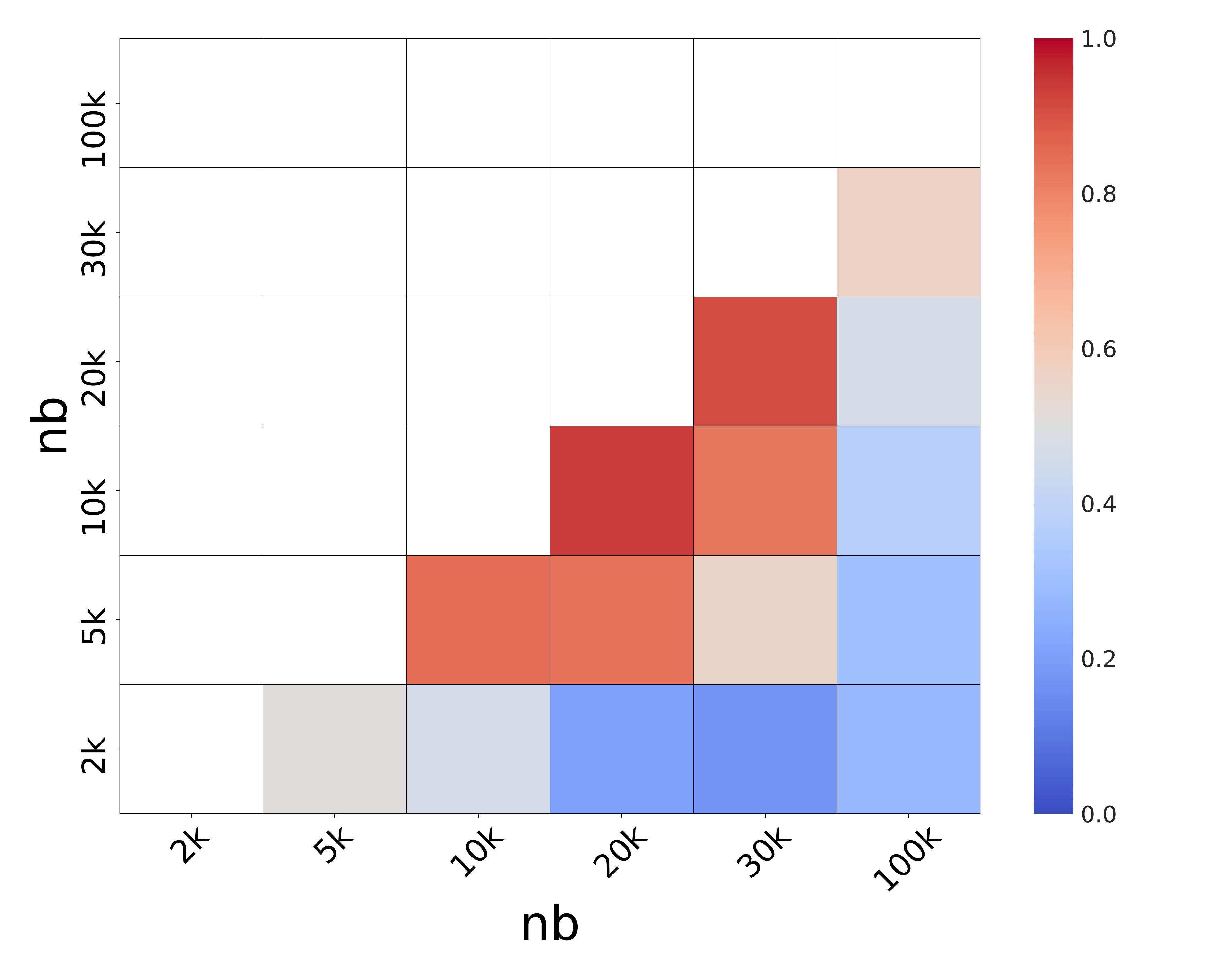}
    \includegraphics[scale=0.09]{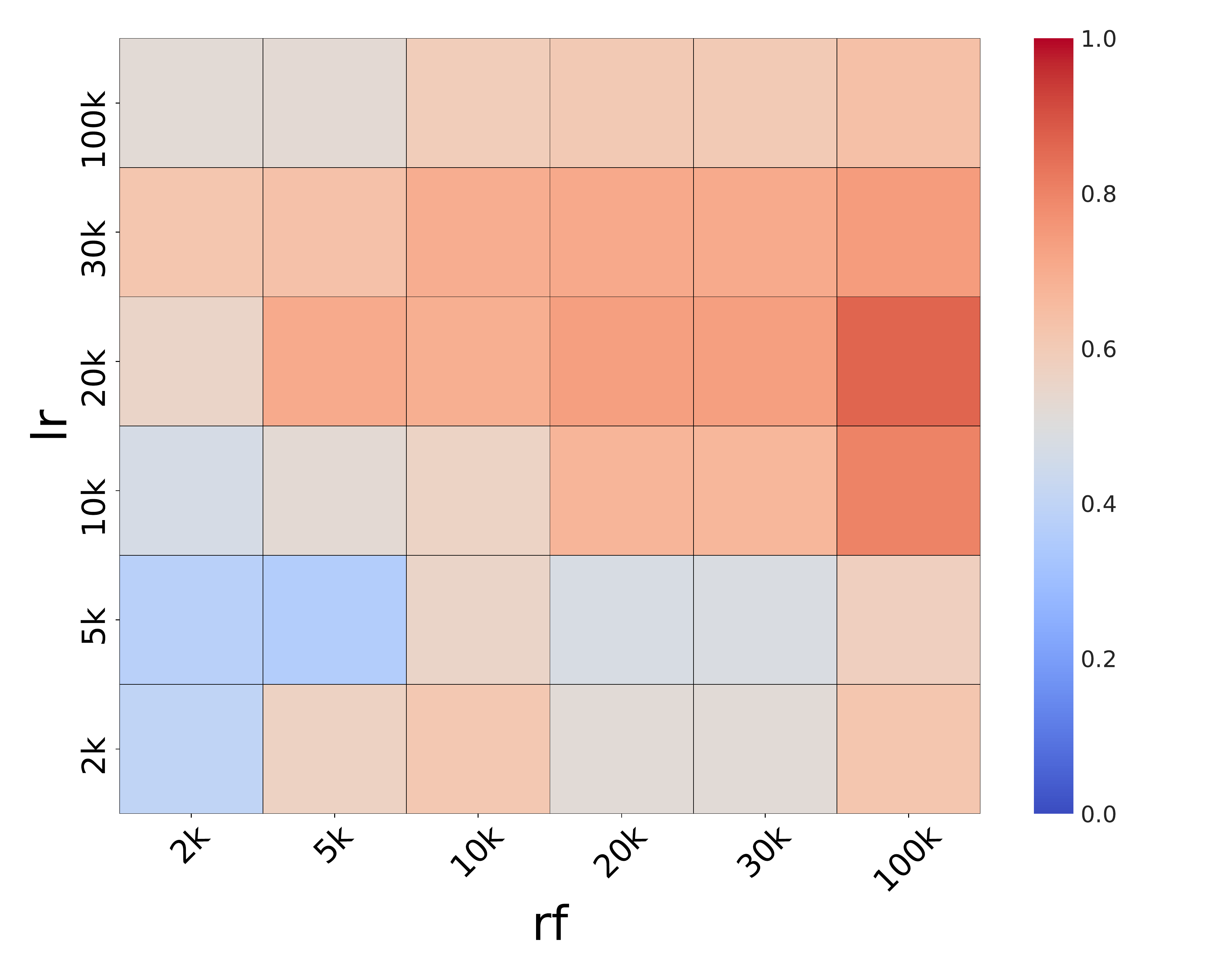}
    \includegraphics[scale=0.09]{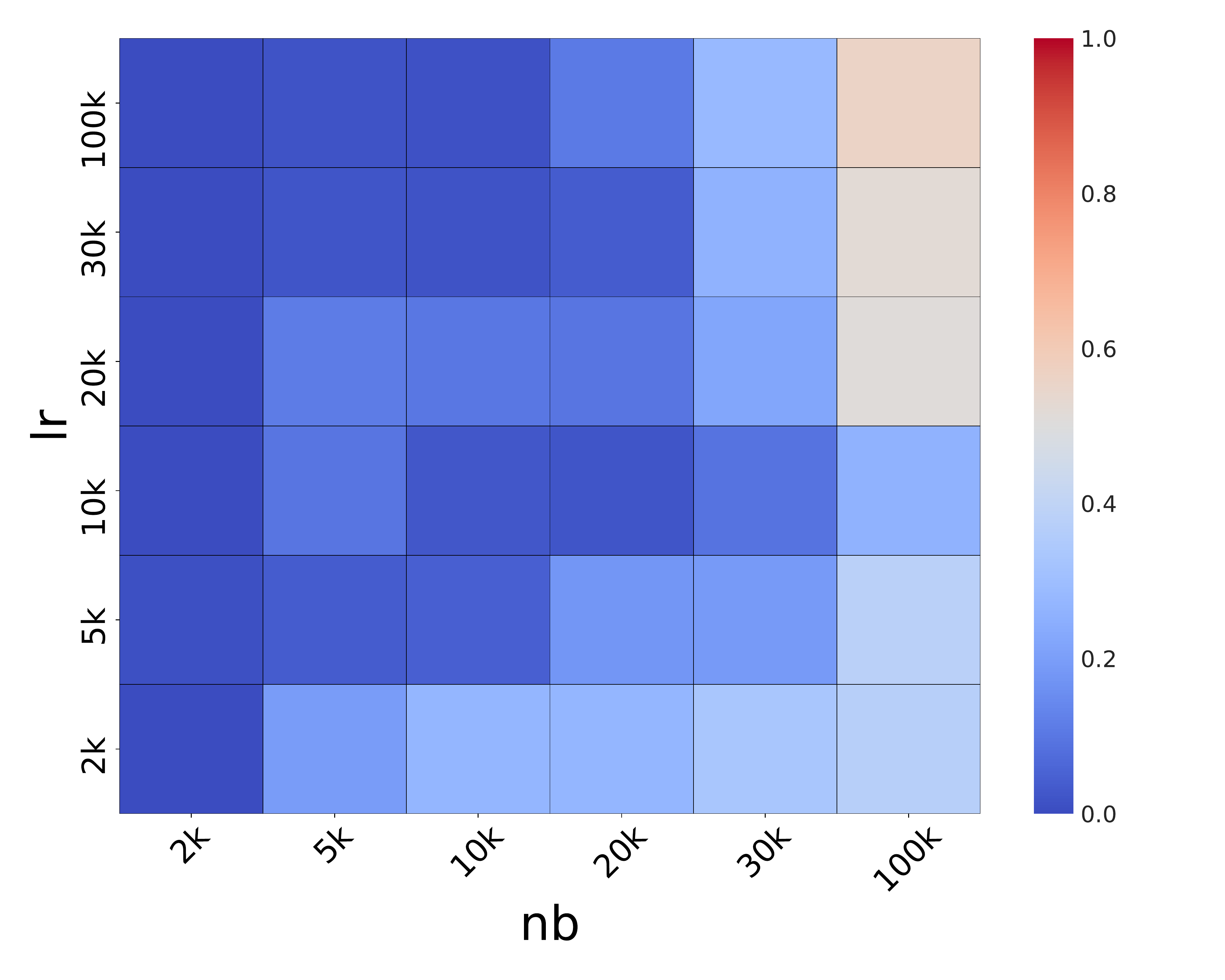}
    \caption{Top: Average correlations sim$_c(s,t)$ for LR (left) and NB (right), using Spearman. Bottom: Average correlations sim$_{c,d}(s,t)$ for $c=$ LR and $d=$ RF (left) and $c=$ LR and $d=$ NB (right).}
    \label{figure:nb_lr_spearman}
\end{figure}

\begin{table}[!htb]
    \centering
    { \small
    \begin{tabular}{l|cccc}
    \toprule
             & LR & RF & NB \\ \midrule
         MLP &  .531/.777 & .312/.564& -.071/.205 \\
         LR &  & .362/.602 & -.107/.147\\
         RF &  & & -.111/.287\\
         \bottomrule
    \end{tabular}
    }
    \caption{Min/Avg values sim$_{c,d}(s,t)$ across $(s,t)$ (using Spearman) between classifiers $c$ and $d$.}
    \label{table:cross_correlations_classifiers_spearman}
\end{table}

\begin{figure}[!htb]
    \centering
    \includegraphics[scale=0.15]{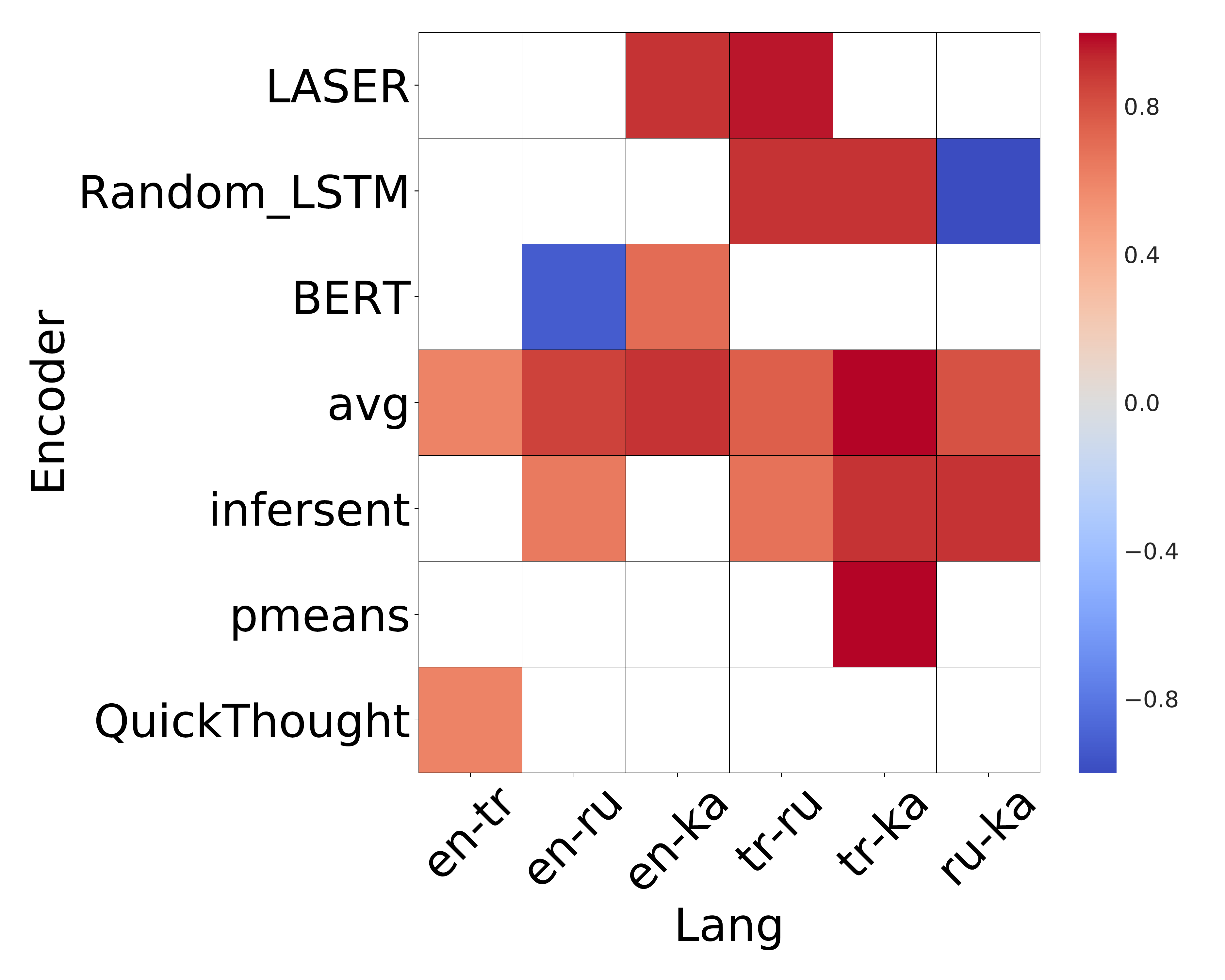}
    \caption{Spearman correlations 
    across languages for different encoders.}
    \label{figure:encoder_lang_spear}
\end{figure}

\begin{figure}[!htb]
    \centering
    \includegraphics[scale=0.15]{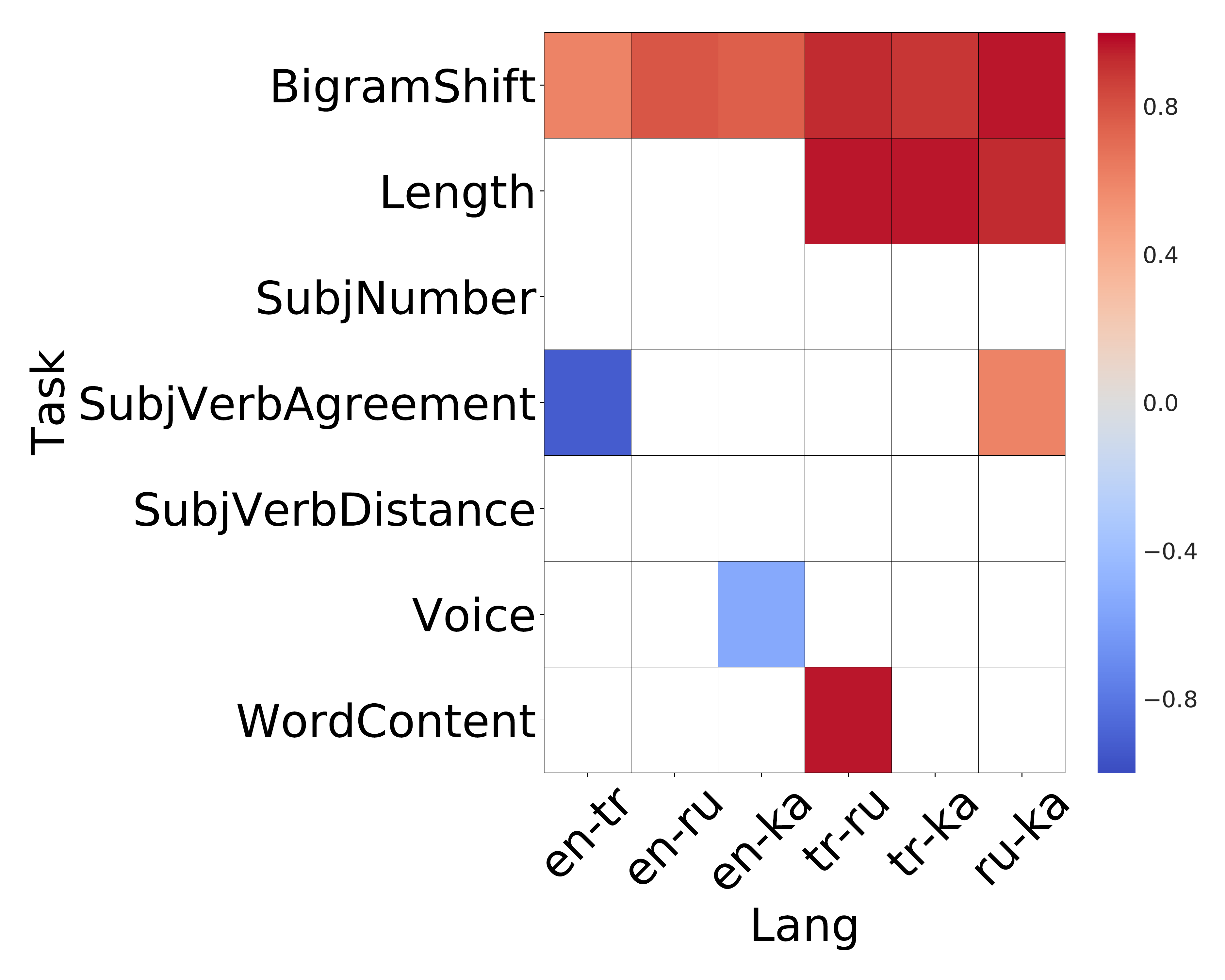}
    \caption{Spearman correlations 
    across languages for different probing tasks.}
    \label{figure:task_lang_spear}
\end{figure}

\begin{figure*}[!htb]
    \centering
    \includegraphics[scale=0.1]{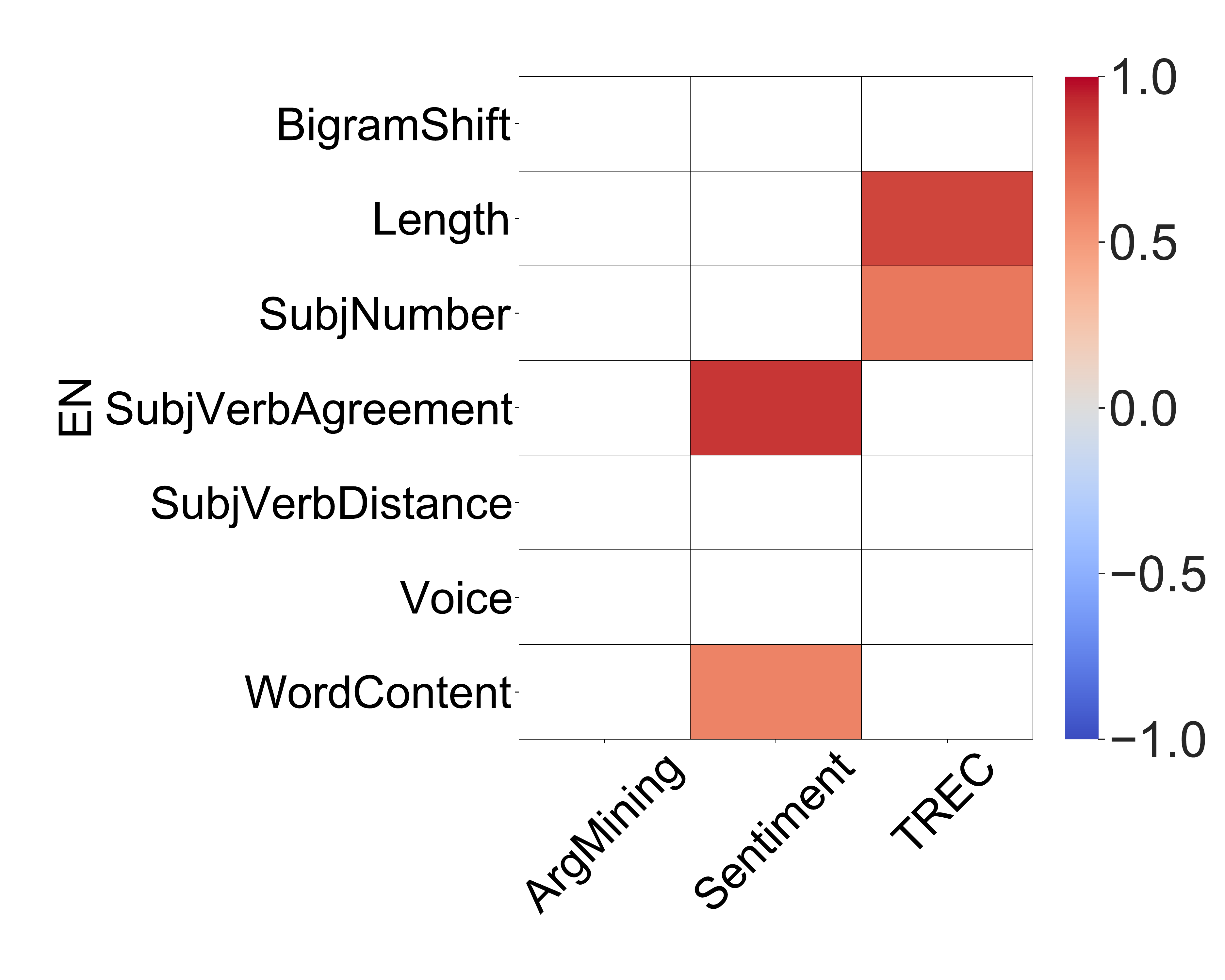}
    \includegraphics[scale=0.1]{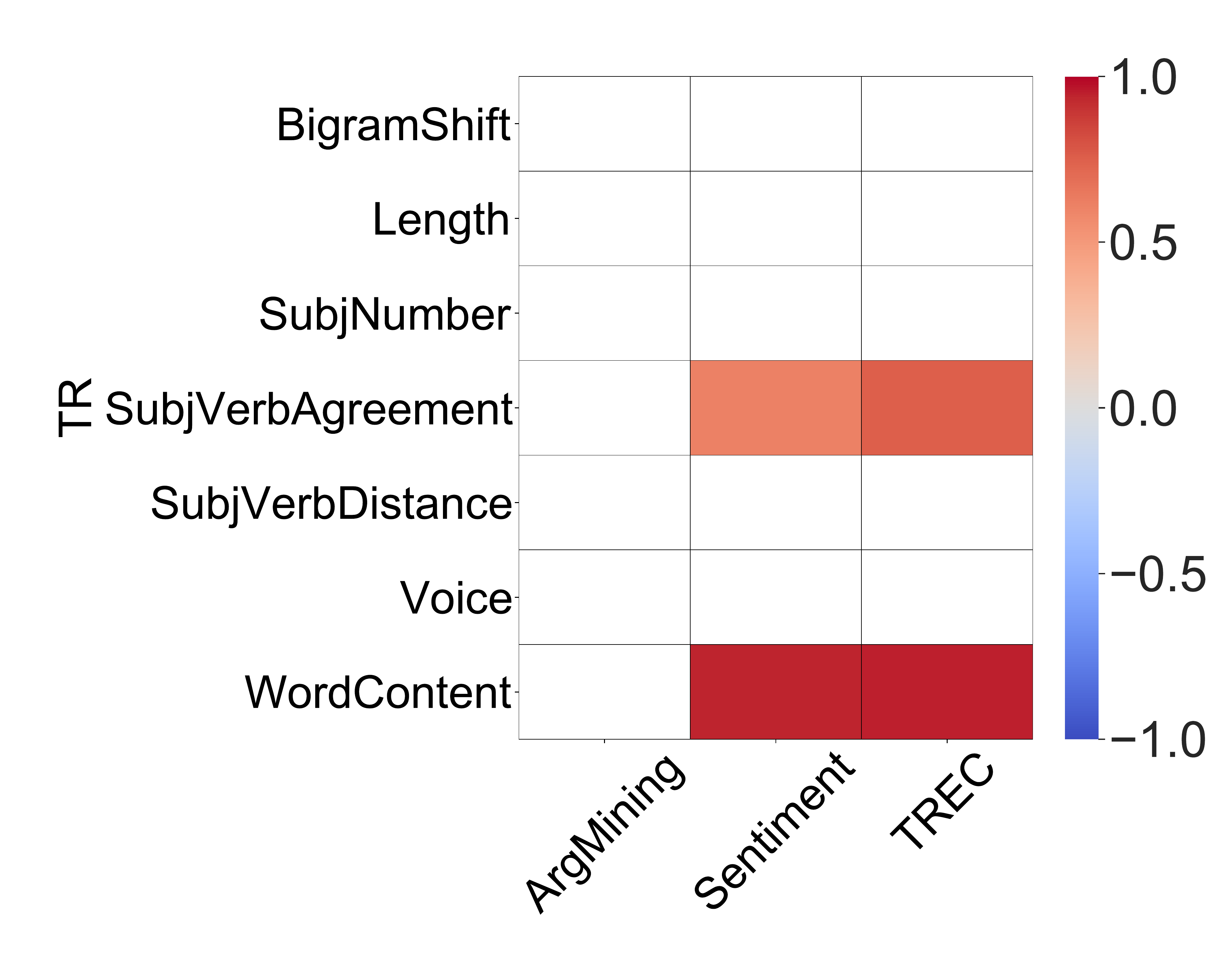}
    \includegraphics[scale=0.1]{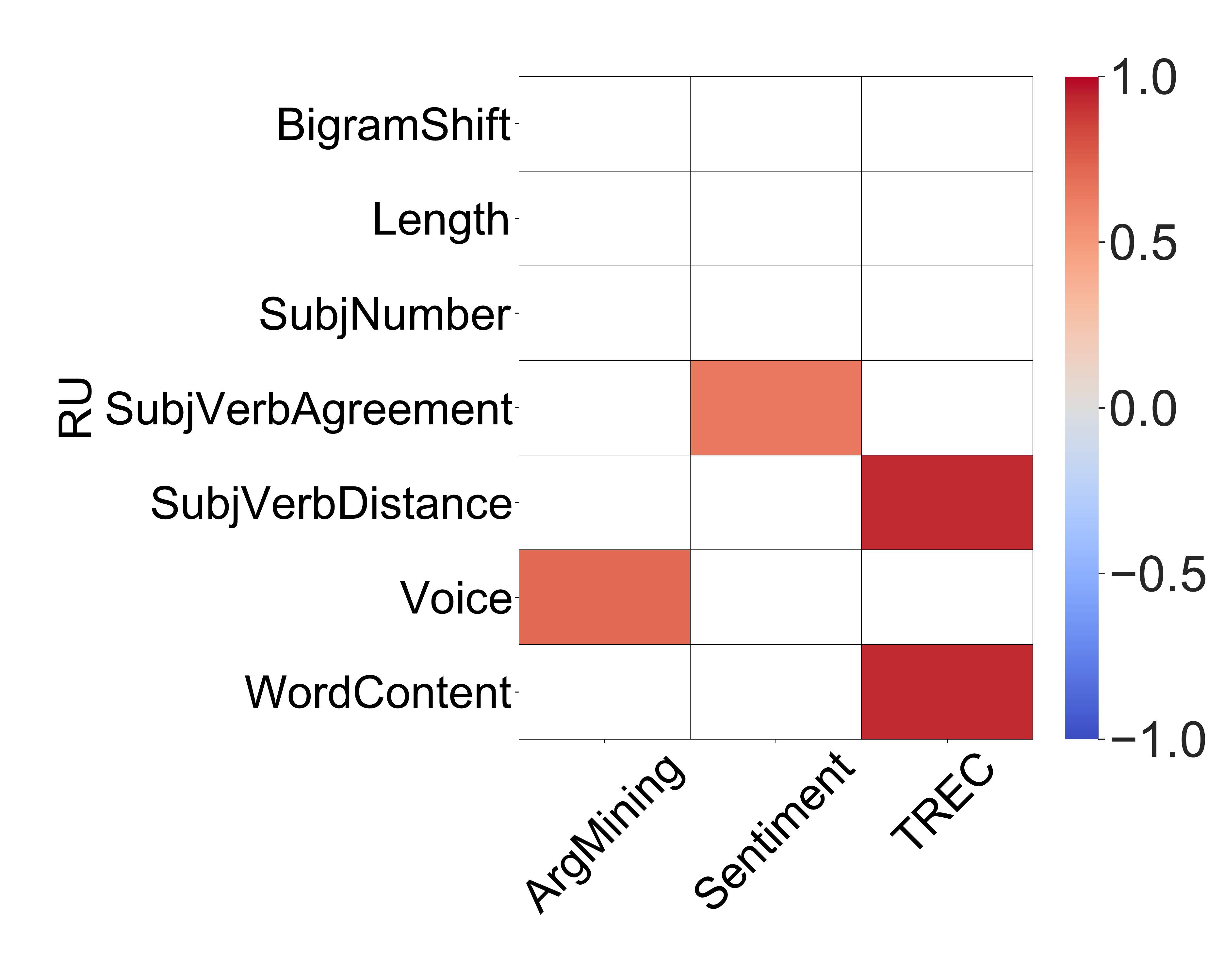}
    \includegraphics[scale=0.1]{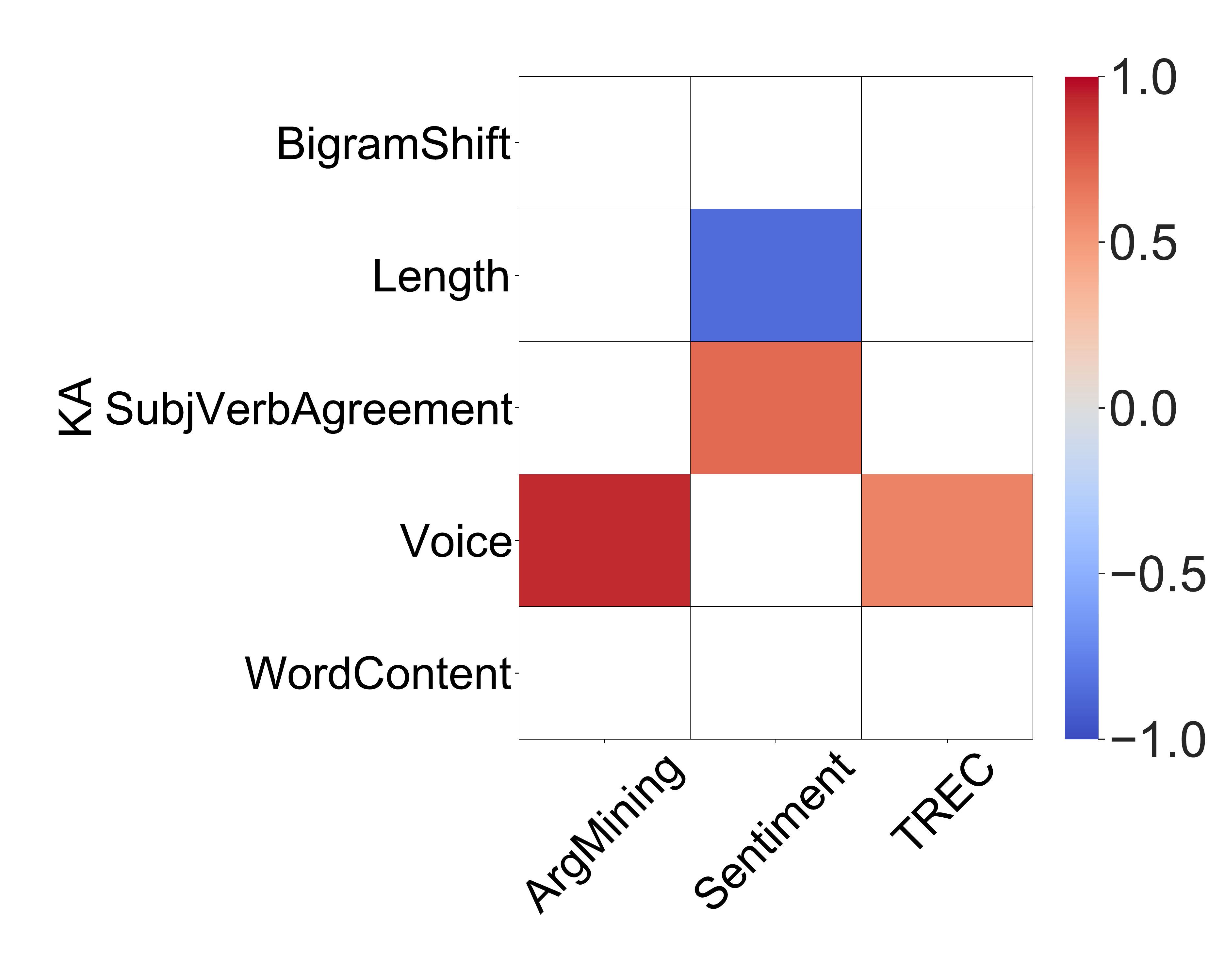}
    \caption{Spearman correlation among probing task and downstream performance for all languages.}
    \label{figure:downstream_multi_spearman}
\end{figure*}

\section{Class Imbalance} 
In addition to the classifier type and size, we also tested the influence of the class (im)balance of the training data.
In particular, for the four binary probing tasks BigramShift, SubjNumber, SV-Agree, and Voice, we examine the effect of imbalancing with ratios of 1:5 and 1:10.
We use LR with sizes of 10k, 20k, and 30k training instances and correlate 
the results for imbalanced datasets with the standardly balanced datasets. We find that (i) for two tasks (BigramShift, SV-Agree) there is typically high correlation (0.6-0.8) while for the other two tasks the correlation is typically zero between the balanced and imbalanced setting; (ii) correlation to the setting 1:1 (slightly) diminishes as we increase the class imbalance from 1:5 to 1:10. Thus, the scenarios 1:5 and 1:10 do not strongly correlate with 1:1 (as used in all our other experiments).
As a consequence, in the multilingual setup, we paid attention to keep datasets as uniform as possible. 

\section{Spearman correlations}

Figures~\ref{figure:nb_lr_spearman}--\ref{figure:downstream_multi_spearman} and Table~\ref{table:cross_correlations_classifiers_spearman} show Spearman correlation results for the corresponding Pearson results in the main text.

\end{document}